\def\year{2019}\relax
\documentclass[letterpaper]{article} %DO NOT CHANGE THIS
\usepackage{aaai19}  %Required
\usepackage{times}  %Required
\usepackage{helvet}  %Required
\usepackage{courier}  %Required
\usepackage{url}  %Required
\usepackage{graphicx}  %Required
\usepackage{amsmath,amssymb,amsfonts}
\usepackage[ruled,linesnumbered]{algorithm2e}
\usepackage{subfigure}
\usepackage{booktabs}

\begin{document}
% The file aaai.sty is the style file for AAAI Press
% proceedings, working notes, and technical reports.
%
\title{Hypergraph Optimization for Multi-structural Geometric Model Fitting}
\author{
Shuyuan Lin$^1$,
Guobao Xiao$^2$,
Yan Yan$^1$,
David Suter$^3$,
Hanzi Wang$^{1*}$ \\
%\\
$^1$ Fujian Key Laboratory of Sensing and Computing for Smart City, \\ School of Information Science and Engineering, Xiamen University, China \\
$^2$ Fujian Provincial Key Laboratory of Information Processing and Intelligent Control, \\College of Computer and Control Engineering, Minjiang University, China\\
$^3$ School of Science, Edith Cowan University, Australia  \\
swin.shuyuan.lin@gmail.com, gbx@mju.edu.cn, \{yanyan, hanzi.wang\}@xmu.edu.cn, d.suter@ecu.edu.au \\
}
\maketitle
\let\thefootnote\relax\footnotetext{$^*$Corresponding author.}
%\footnote{The corresponding author.}
\begin{abstract}
Recently, some hypergraph-based methods have been proposed to deal with the problem of model fitting in computer vision, mainly due to the superior capability of hypergraph to represent the complex relationship between data points.
However, a hypergraph becomes extremely complicated when the input data include a large number of data points (usually contaminated with noises and outliers), which will significantly increase the computational burden.
In order to overcome the above problem, we propose a novel hypergraph optimization based model fitting (HOMF) method to construct a simple but effective hypergraph.
Specifically, HOMF includes two main parts: an adaptive inlier estimation algorithm for vertex optimization and an iterative hyperedge optimization algorithm for hyperedge optimization.
The proposed method is highly efficient, and it can obtain accurate model fitting results within a few iterations. 
Moreover, HOMF can then directly apply spectral clustering, to achieve good fitting performance.
Extensive experimental results show that HOMF outperforms several state-of-the-art model fitting methods on both synthetic data and real images, especially in sampling efficiency and in handling data with severe outliers.
\end{abstract}
\section{Introduction}
\label{sec:introduction}
Robust fitting of geometric structures for multi-structural data contaminated with a number of outliers is one of the most important and challenging research tasks for many applications of computer vision \cite{FischlerBolles1981}, such as 2D line fitting \cite{li2009consensus}, vanishing point detection \cite{tardif2009non}, two-view segmentation \cite{magri2014t} and 3D-motion segmentation \cite{ochs2012higher}.
The task of robust model fitting is to accurately and efficiently recover meaningful structures from data.
However, data in many applications are often contaminated with noises and outliers, which makes the problem of model fitting challenging.
\begin{figure*}[!t]
	\centering
	\includegraphics[width=1\textwidth]{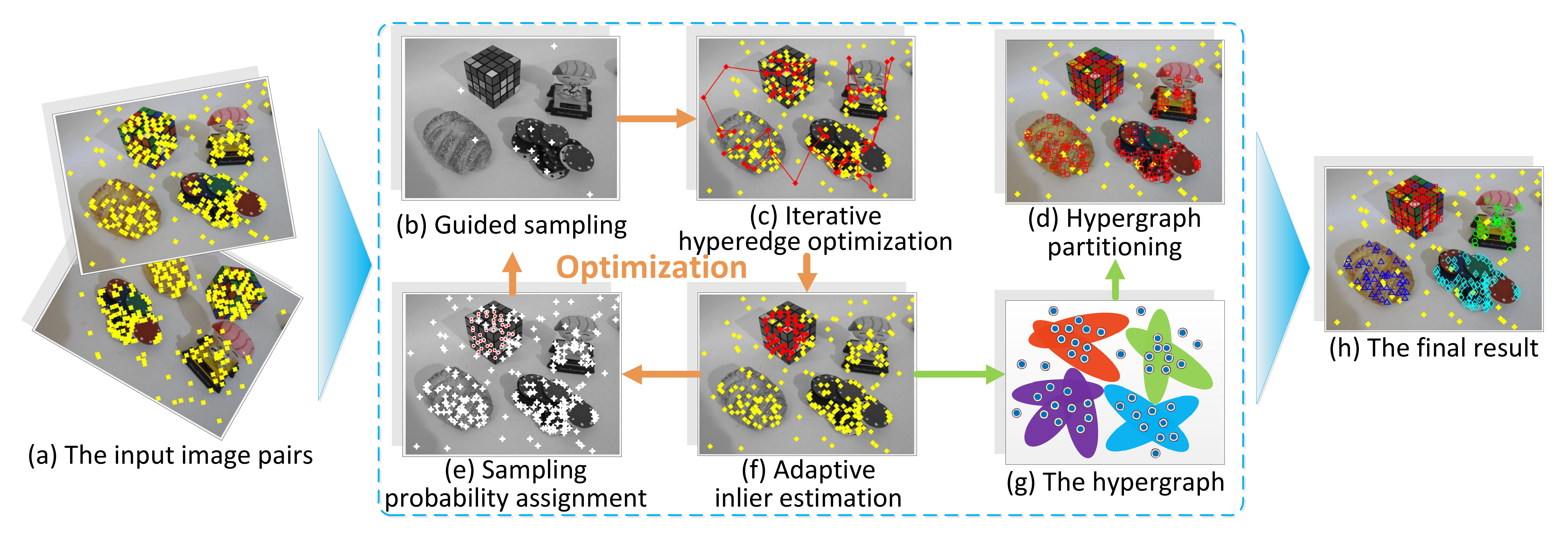}
	\caption{The overview of the proposed framework for two-view motion segmentation. (a) The outliers and the inliers are denoted in the yellow color. (b) Sampling points are marked with the white color. (c), (d) and (f) The sampled points are denoted in the red color and the other points are denoted in the yellow color. The red curves are denoted as the hyperedges. (e) Sampling points are marked with the white color and the red points surrounded by the white color means that they are assigned with a lower sampling probability. (g) The hypergraph with four hyperedges and some vertices. (h) The inliers belonging to different structures are denoted in the red, green, cyan and blue colors, respectively. The outliers are denoted in the yellow color.}\label{fig:framework}
\end{figure*}	
\par
In the past few decades, the hypergraph representation has attracted much attention in many computer vision applications \cite{parag2011supervised,jain2013efficient,huang2016regression}.
A hypergraph modelling is an extended version of the traditional graph modelling.
A graph modelling can be used to describe data through vertices and edges (an edge of a graph connects only two vertices), and a pairwise similarity measure \cite{saito2018hypergraph}.
Compared with a graph modelling, a hypergraph modelling can be used to effectively describe complex data relationship based on hyperedges (each of which may connect more than two vertices).
For example, a probabilistic hypergraph method proposed by Huang et al. \cite{huang2010image} establishes the relationship between the vertices and the hyperedge in terms of the local grouping information and the similarity in a probabilistic way.
Generally speaking, the hypergraph modelling not only inherits the basic properties of the graph modelling, but also shows superior advantages over the graph modelling.
\par
Recently, hypergraph analysis has been successfully applied to robust model fitting and it has achieved promising performance \cite{zhou2007learning,agarwal2005beyond}.
The traditional hypergraph analysis usually considers the relationship between  hyperedges and vertices.
However, since data in practical tasks are often contaminated with noises and outliers, the traditional hypergraph-based methods suffer from two issues: 1) the hypergraph construction becomes quite complex when data points are contaminated with many outliers, and 2) a large number of hyperedges generating from noisy vertices generally increase the computational cost.
Therefore, it is quite important to optimize hypergraphs to reduce the computational complexity, which has not been well studied yet.
\par
In this work, we propose a novel hypergraph optimization method (i.e., HOMF) for robust model fitting (as shown in Fig. \ref{fig:framework}) to overcome the above problems.
HOMF can not only fit multi-structural data contaminated with both noises and outliers, but also effectively reduce the computational complexity. The main contributions of this paper are summarized as follows:
\begin{itemize}
\item We present an adaptive inlier estimation algorithm (AIE) based on the weighting scores of data points. As a result, AIE can effectively distinguish significant data points (i.e., inliers) from insignificant data points (i.e., outliers).
\item We develop an iterative hyperedge optimization algorithm (IHO) to accelerate the optimization of hyperedges. IHO can quickly generate the optimized hyperedges with only a few iterations based on an effective `exiting criterion', which is satisfied when the samples are selected from the same structure.
\item We propose a novel hypergraph optimization method (HOMF) by taking advantage of IHO and AIE, which can be used for the guided sampling of different structures. Therefore, the proposed method is highly efficient and can efficiently obtain competitive fitting results.
\end{itemize}
\par
The rest of the paper is organized as follows.
Firstly, we review the related work.
Then, we propose an adaptive inlier estimation algorithm and an iterative hyperedge optimization algorithm for hypergraph optimization.
Next, we report the experimental results obtained by our method and by several competing methods, on both synthetic data and real images.
Lastly, we draw conclusions.
\section{Related Work}\label{sec:Related Work}
As the proposed method is closely related to scale estimation, guided sampling and hypergraph modelling, in this section, we briefly review work related to these.
\par
Robust scale estimation plays a critical role for model fitting.
A number of robust scale estimation methods (e.g., \cite{wang2012simultaneously,litman2015inverting,tiwari2016robust}) have been proposed for the multi-structural fitting task.
Wang et al. \cite{wang2012simultaneously} propose the Iterative $K$-th Ordered Scale Estimator (IKOSE), which is one of the popular robust scale estimation methods due to its accuracy and efficiency.
However, in practice, IKOSE is sensitive to the $K$-th sorted absolute residual.
Tiwari et al. \cite{tiwari2016robust} present the Density Preference Analysis (DPA) that estimates the scale of inlier noise by using linear extrapolation based residual density profile. But DPA overly relies on preference analysis, which focuses on the preference of data points to different models.
In this paper, we propose a novel robust adaptive scale estimator (i.e., AIE).
The proposed AIE can efficiently estimate the scale of significant data points for heavily corrupted multi-structural data.
As a result, the performance of model fitting can be significantly improved by using AIE.
\par
Guided sampling can accelerate multi-stuctural data search by utilizing meta-information on the data distribution \cite{pham2014random,tennakoon2018effective}.
The Random Cluster Model Simulated Annealing (RCMSA) \cite{pham2014random} guides promising hypothesis generation by constructing a weighted graph in a simulated annealing framework. However, the disadvantage of RCMSA is that it assumes spatial smoothness of the inliers, which is computationally expensive and may not apply to particular situations.
The guided sampling method that is most closely related to ours, cost-based sampling (CBS) \cite{tennakoon2018effective} uses a data sub-sampling strategy to generate the hypotheses.
Specifically, CBS employs a $K$-th order statistical cost function to improve the distribution of hypotheses.
That method, however, relies on prior knowledge and the greedy algorithm.
Unlike these previous works, we use AIE to identify the insignificant data points (i.e., outliers) for guided sampling, which is more efficient for rapidly sampling minimal subsets for different structures.
\par
Recently, some hypergraph-based methods have been proposed in computer vision, e.g., \cite{ochs2012higher,wang2018searching}.
For example, Wang et al. \cite{wang2018searching} propose a mode-seeking algorithm for searching the representative modes on hypergraphs, which is more effective than the conventional mode-seeking methods for model fitting.
Note that, the proposed method focuses on hypergraph optimization, which aims to generate an optimized hypergraph that is more suitable for spectral clustering.
The hypergraph optimization problem can be treated as the combination of the vertex labeling problem and the hyperedge estimation problem.
However, it is challenging to solve both problems simultaneously.
In this paper, we solve the hypergraph optimization through an iterative updating strategy, by which the hypergraph can be optimized step by step during the iterative process. Then, spectral clustering is used for partitioning the optimized hypergraph after hypergraph optimization.
%===============================================
\section{Methodology}\label{sec:Our method}
In this section, we describe the proposed HOMF, which takes advantage of an adaptive inlier estimation algorithm (AIE) and an iterative hyperedge optimization algorithm (IHO), for the geometric model fitting problem.
Specifically, we first develop AIE to select the significant data points (i.e., inliers).
Then we propose IHO for accelerating the optimization of initial hyperedges.
Lastly, we present the complete HOMF method.
%===================================================
\subsection{Adaptive Inlier Estimation}
\label{ssec:Adaptive_inliers_estimation_for_hypergraph}
The inlier scale estimation plays a critical role in the hypergraph optimization. However, most of the scale estimators need to manually choose a threshold for determining the number of model instances \cite{wang2012simultaneously}.
To address the above problem, we propose AIE to adaptively estimate the inlier noise scale.
Specifically, AIE refines IKOSE \cite{wang2012simultaneously} and kernel density estimation (KDE) \cite{silverman1986density} to compute the weighting score of each data point $x_i$ for a generated model hypothesis $h$ using the following equation:
	\begin{equation}
		\label{equ:w(v)}
		\omega_i = \frac{1}{nb} \left[ \mathbf{EK}\left(r^h_i/b\right)\right]
	\end{equation}
where $\mathbf{EK}(\cdot)$ is the popular Epanechnikov kernel function \cite{wand1995mc}; $\mathbf{r}^h=\{r_i^h\}_{i=1}^{n}$ is the residual set between each data point $x_i$ and the model hypothesis $h$;
$n$ is the number of data points. $b$ is a bandwidth defined as follows \cite{wand1995mc}:
	\begin{equation}
		\label{equ:b(v)}
		b = \left[ \frac{7\int_{-1}^{1}\mathbf{EK}\left(\mathbf{r}^h\right)^2dr}{n\int_{-1}^{1}(\mathbf{r}^h)^2\mathbf{EK}\left(\mathbf{r}^h\right)dr} \right]^{0.2}
	\end{equation}
Inspired by \cite{sezgin2004survey,ferraz2007density}, we select significant data points by using a simple but effective data driven thresholding technique. Given a set of data points $X=\{x_{1}, x_{2}, ..., x_{n}\}$ and the corresponding squared weighting scores $\mathbf{w}^2=\{\omega_1^2, \omega_2^2, ..., \omega_n^2\}$, we define $\xi_i = \mathrm{max}\{\mathbf{w^2}\}-\omega_i^2$, which denotes the gap between the squared weighting score of data points $X$ and the squared weighting score of the data point $x_i$ for the given model hypothesis $h$.
Note that the logarithm is not meaningful when the gap $\xi_i$ is a negative value. The prior probability $p(\xi_i)$ can be computed as $p(\xi_i) = \xi_i \big/  \sum_{j=1}^{n}\xi_j$.
\par
The entropy of the prior probability for all data points can be computed as follows:
	\begin{equation}
			\label{equ:entropy}
			\Pi=-\sum_{i=1}^{n}p(\xi_i)\log p(\xi_i)
	\end{equation}
\par
The entropy is chosen as the threshold to distinguish the significant data points from the insignificant data points, as follows:
	\begin{equation}
			\label{equ:significant}
			\mathbf{\vartheta}^* =  \bigg\{x_i \bigg | -\log p(\xi_i) > \Pi \bigg\}
    \end{equation}
Here, we use information theory \cite{wang2012simultaneously} in Eq. (\ref{equ:significant}) to select the significant data points and reject the other insignificant data points.
It is worth pointing out that the difference between \cite{wang2012simultaneously} and the proposed is that AIE can adaptively choose the number of significant data points independent of the $K$-th sorted absolute residual.
%=======================================
\begin{algorithm}[!t]
	\caption{The adaptive inlier estimation (AIE)}
	\label{algorithm1}
	\KwIn{The residuals (to a hypothesis) $\boldsymbol{r}^h$ and the number of data points $n$.}
	\KwOut{The significant data points $\boldsymbol{\vartheta}^*$.}
	Compute the bandwidth $b$ of the kernel density function by Eq. \eqref{equ:b(v)}.\\
	Estimate the weighting score for each data point by Eq. \eqref{equ:w(v)}. \\
	Calculate the entropy $\Pi$ based on the set of the weighting scores by Eq. \eqref{equ:entropy}. \\
	Select the significant data points (i.e., inliers) $\boldsymbol{\vartheta}^*$ by Eq. \eqref{equ:significant}. \\
\end{algorithm}
%=======================================
%=======================================
\begin{algorithm}[!b]
	\caption{The iterative hyperedge optimization (IHO)}
	\label{algorithm2}
	\KwIn{The initial hyperedge $\mathcal{E}(e)$, the vertices $\mathcal{V} = {\{v_i\}}_{i=1}^{n}$, the minimum tolerable size $q$, the higher than minimal subset $l$ and the number of iterations $T_{max}$.}
	\KwOut{The optimized hyperedge $\mathcal{\hat{E}}(e_t)$.}
	\For{t=$1$ to $T_{max}$}{
		Calculate the residual $\mathbf{r}^h$ between the hyperedge and the vertices. \\
		Estimate the weighting score $\omega$ of each element in $\mathbf{r}^h$ by Eq. \eqref{equ:w(v)} to obtain the weighting score set $\mathbf{w}_{t}$.\\
		Sort $\mathbf{w}_{t}$ in the ascending order to obtain the permutation $\tilde{\mathbf{w}}_{t}$. \\
		Generate a new hyperedge $\mathcal{\hat{E}}(e_t)$ by refitting the vertices corresponding to the $[\tilde{\mathbf{w}}_{t}]_{q-l}^{q}$.  \\
		Evaluate $Q_{e}$ in Eq. \eqref{equ:exitingcriterion}. \\
		\lIf{$Q_{e} == 1$}{break}
	}
\end{algorithm}
%======================================
\subsection{Hypergraph Optimization for Model Fitting}\label{sec:Fasthypergraphconstruction}
In this paper, a hypergraph is defined as ${G=(\mathcal{V}, \mathcal{E}, \mathcal{W})}$, which includes the vertex set ${\mathcal{V}=\{v_{1}, v_{2}, ..., v_{n}\}}$, the hyperedge set ${\mathcal{E}=\{e_{1}, e_{2}, ..., e_{m}\}}$, and the weight set $\mathcal{W}=\{\omega_{1}, \omega_{2}, ..., \omega_{n}\}$, where $n$ and $m$ are respectively the number of vertices and the number of the hyperedges.
Each vertex is assigned a weighting score $\omega_i$ (see Sec. Adaptive Inlier Estimation).
The hypergraph modelling $G$ is an extension of an ordinary graph modelling, where the hyperedge $e$ might connect more than two vertices, to also include weights.
\par
In our hypergraph construction, each vertex is defined as a data point and each hyperedge corresponds to a model hypothesis.
We construct the hyperedge of hypergraph based on the higher than minimal subset sampling (HMSS) algorithm due to its good accuracy and efficiency, which has been demonstrated in \cite{tennakoon2016robust}.
\par
Similar to \cite{govindu2005tensor,ochs2012higher,tennakoon2018effective}, in order to make the hypergraph optimization tractable, we decompose the hypergraph by multiplying the pairwise affinity matrix $\mathbf{H}$ with its transpose.
Each column of the affinity matrix $\mathbf{H}$, which is obtained by encoding more than two vertices as a subset, corresponds to the hyperedge. The affinity matrix $\mathbf{H}$ characterizes the relationships between hyperedges $\mathcal{E}$ and vertices $\mathcal{V}$.
Specifically, the simple version of the hypergraph $G^*$ is represented as follows:
	\begin{equation}
		\label{equ:g=hh}
		%   G^* = \sum_{i}{H_iH_i^\mathsf{T}}
		G^* = \mathbf{HH}^\mathsf{T} =\sum_{i}^{m}{\bigg[exp^{-\frac{\mathbf{w}_{i}}{2\sigma^2}}\bigg]\bigg[exp^{-\frac{\mathbf{w}_{i}}{2\sigma^2}}\bigg]^\mathsf{T}}
\end{equation}
where $exp^{(\cdot)}$ is the exponential map. $\mathbf{w}_{i}$ is the set of weighting scores, which is computed by Eq. (\ref{equ:w(v)}) between the hyperedge $e_i$ and vertices $\mathcal{V}$. $\sigma$ is a normalization constant. $[\cdot]^\mathsf{T}$ is the transpose of $[\cdot]$. $m$ is the number of hyperedges.
The hypergraph $G^*$ contains many redundant vertices and hyperedges, which will lead to the high computational complexity. Thus, it needs to be optimized step by step during the subsequent iterative updating process (see Alg. \ref{algorithm2}).
The detail of the hypergraph optimization is given as follows:
%=====================================
\begin{algorithm}[!t]
	\caption{The hypergraph optimization based model fitting (HOMF) method}
	\label{algorithm3}
	\KwIn{A set of data points $X=\{x_i\}_{i=1}^{n}$, the number of model hypotheses $m$ and the number of structures $c$.}
	\KwOut{The model instances and the hyperedges.}
	Initialize the sampling probability $P=\{\mathbf{\rho}_i\}_{i=1}^{n}$ of all data points to 1. \\
	Generate a model hypothesis $h$ by random sampling. \\
	Construct a hypergraph $G^*$ based on the generated model hypothesis $h$. \\
	\For{j=$1$ to $m$}{
		\If{j \textgreater $1$ }
		{
			Generate a new model hypothesis $h_j$ according to the sampling probability $P$ and add a hyperedge $\mathcal{E}(e_j)$ according to $h_j$ in $G^*$. \\
		}
		Generate an optimized hyperedge $\mathcal{\hat{E}}(e_j)$ by Alg. \ref{algorithm2}.\\
		Calculate the residual $\boldsymbol{r}^h_j$ between the vertices and optimized hyperedge $\mathcal{\hat{E}}(e_j)$.\\
		Optimize the vertices connected to the optimized hyperedge $\mathcal{\hat{E}}(e_j)$ using AIE by Alg. \ref{algorithm1}.\\
		Update the sampling probability of the vertices $P$. \\
	}
	Segment the hypergraph by spectral clustering to obtain the model instances and the hyperedges. \\
\end{algorithm}
%============================================
\par
Firstly, we generate a new model hypothesis using the random sampling with HMSS \cite{tennakoon2016robust} to add an initial hyperedge in the hypergraph.
\par
Then, we estimate the set of weighting scores $\mathbf{w}_{i}$ that will be sorted as an ordered weighting scores permutation $\tilde{\mathbf{w}}_{i}$.
\par
Thirdly, we adopt an iterative algorithm to effectively and efficiently optimize the hypergraph. Similar to \cite{tennakoon2016robust}, we determine whether the hypergraph construction process converges based on the last three iterations. However, the difference between the proposed method and \cite{tennakoon2016robust} is that we use the weighting scores of the vertices as condition for the `exiting criterion', whose advantage is that it can reduce the sensitivity to residuals.
In contrast, \cite{tennakoon2016robust} uses the residual of data points as conditions.
Specifically, the `exiting criterion' $Q_{e}$ is formulated as follows:
	\begin{equation}
			\label{equ:exitingcriterion}
			Q_{e} = \bigg ( \omega_{q, t}^{2} \small{<} \sum_{j=q-l}^{q} \omega_{j, t-1}^{2} \bigg ) \small{\wedge} \bigg ( \omega_{q, t}^{2} \small{<} \sum_{j=q-l}^{q} \omega_{j, t-2}^{2} \bigg )
	\end{equation}
where $l$ and $q$ are respectively the higher than minimal subset (i.e., the minimal subset $p+2$) and the minimum tolerable size of the same structure ($q \gg l$, in our experiment, we set the $q$ to be $0.1 \times n$).
$t$ denotes the number of current iteration.
$\omega_{q, t}$ is the weighting score in current iteration $t$ with respect to the minimum tolerable size $q$.
The results obtained from the last three iterations (i.e., $t$, $t-1$ and $t-2$ iterations) are more likely to belong to the same structure  \cite{tennakoon2016robust}.
The above steps can quickly generate the optimized hyperedge but cannot effectively remove some redundant vertices (i.e., insignificant data points), which will affect the following sampling procedure (guided sampling).
Hence, we use AIE (see Sec. Adaptive Inlier Estimation) for estimating the vertices (corresponding to the significant data points) of optimized hyperedge to solve this problem.
\par
After that, the weighting scores of vertices corresponding to the significant data points are selected to optimize the affinity matrix $\mathbf{H}$ and the vertices corresponding to the insignificant data points are assigned a higher sampling probability. 
Specifically, the sampling probability of the significant vertices is gradually  increased about 2-10 times, while the sampling probability of the other vertices is gradually reduced about 2-10 times during each update process.
\par
Finally, these vertices (corresponding to the insignificant data points) will guide the following sampling procedure, which will focus on sampling for different structures.
This way can effectively improve the contribution of vertices corresponding to the insignificant data points for hypergraph optimization.
\par
After the hypergraph optimization, the vertices of each hyperedge can be extracted as the optimized column of the affinity matrix $\mathbf{H}$, so that we only need to deal with a small hypergraph.
At the same time, it allows us to directly apply the spectral clustering-based algorithm to obtain the final segmentation results.
%============================================
\subsection{The Complete Method}\label{sec:The complete method}
In the previous subsections, we gave all the components of the proposed HOMF method. Now we describe the complete algorithm in Alg. \ref{algorithm3}.
Firstly, random sampling is used for generating an initial hypergraph.
Then, IHO is employed to accelerate the optimization of hyperedges.
The vertices (corresponding to the significant data points) of the optimized hyperedges are then estimated using AIE.
Secondly, we assign a lower sampling probability for each vertex (corresponding to the significant data point) and a higher sampling probability for other vertices (corresponding to the insignificant data points).
In this way, the vertices corresponding to the significant data points can be selected to reduce the computational complexity and the vertices corresponding to the insignificant data points can be used for guided sampling, which in turn enhances hypergraph construction.
This optimization process is performed iteratively until the `exiting criterion' is reached (or the fixed number of iterations is reached).
Fortunately, the `exiting criterion' causes the proposed method only perform only a few iterations.
Lastly, spectral clustering is applied to obtain the model instances and the hyperedges.
%=====================================
\section{Experiments}\label{sec:Experiments}
\label{sec:c4}
In this section, we firstly evaluate the performance on synthetic data of AIE compared several robust scale estimation methods including the median (MED), the median absolute deviation (MAD), KOSE \cite{lee1998robust}, IKOSE \cite{wang2012simultaneously}, AIKOSE \cite{wang2013amsac} and DPA \cite{tiwari2016robust}. Then we compare the performance on real images of proposed HOMF with several state-of-the-art model fitting methods including CBS \cite{tennakoon2018effective}, MSHF \cite{wang2018searching}, RPA \cite{magri2015robust}, RCMSA \cite{pham2014random} and UHG \cite{lai2017unified}.
All experimental results are obtained by running 50 times.
%======================================
\begin{figure}[!t]
	\centering
	\subfigure[Two intersecting lines data with 50\% outliers]{\label{Fig.sub.2}
		\includegraphics[width=0.19\textwidth,height=1.1in]{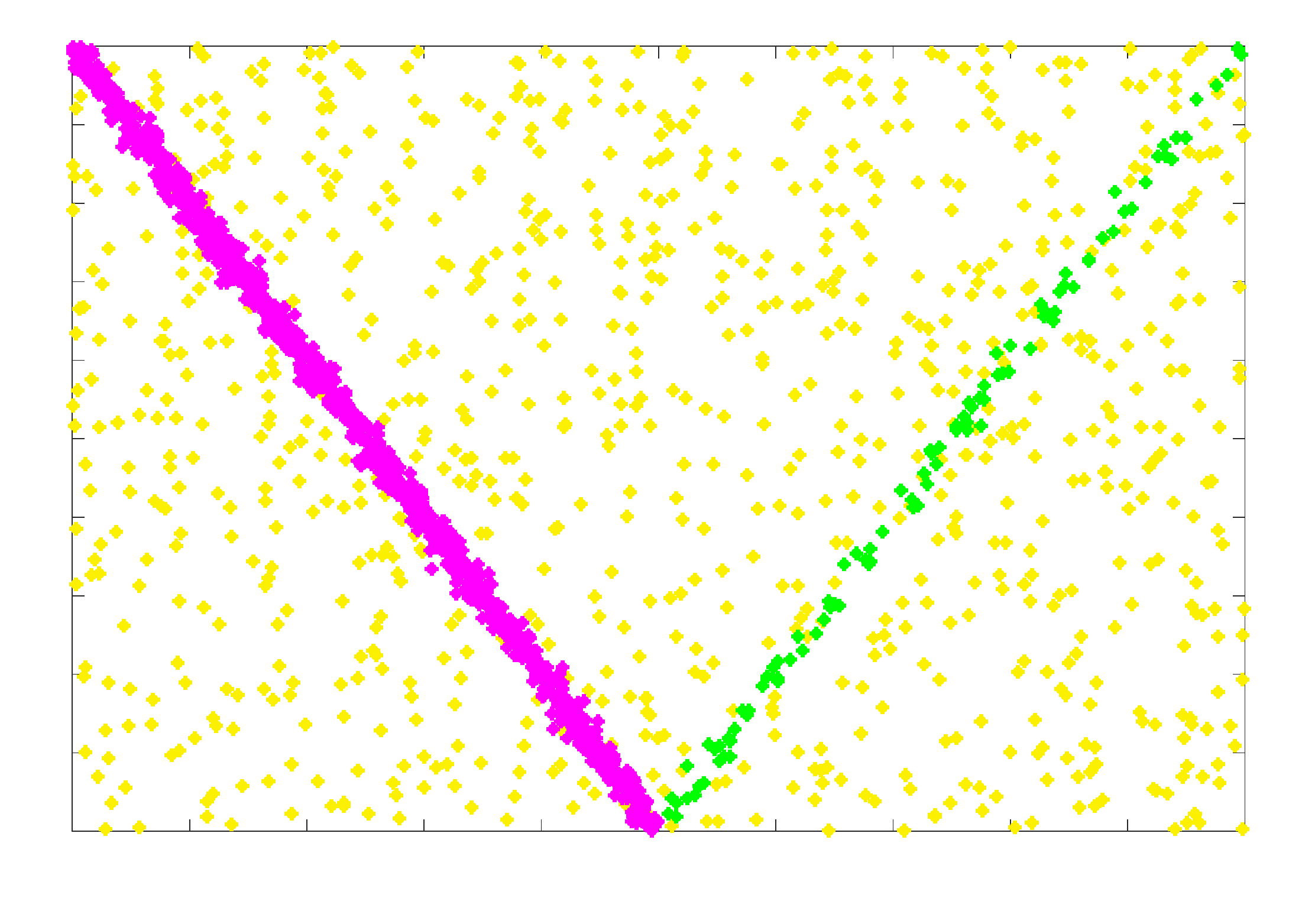}}
	\subfigure[Two intersecting lines data with 95\% outliers]{\label{Fig.sub.3}
		\includegraphics[width=0.19\textwidth,height=1.1in]{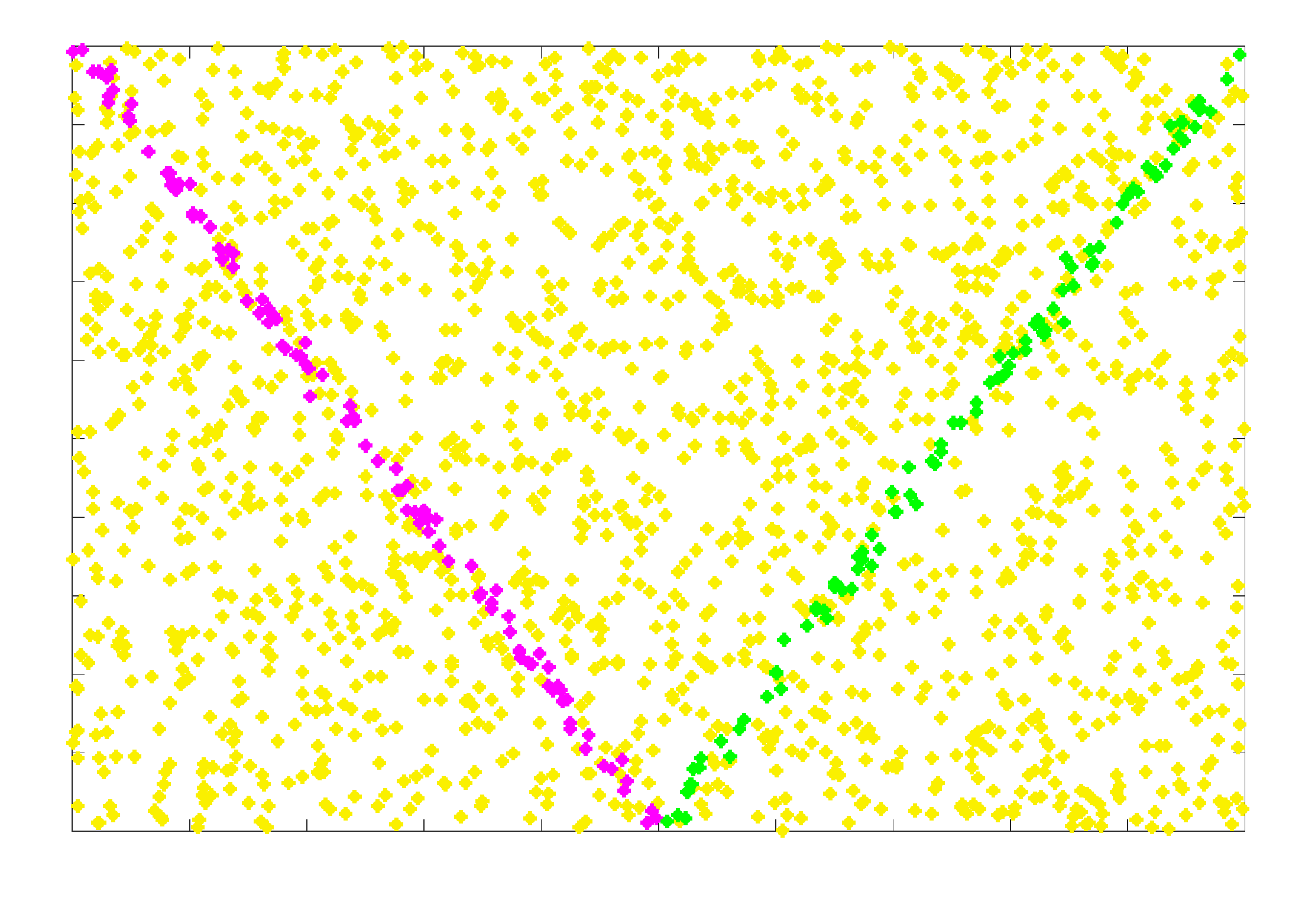}}
	\subfigure[The mean errors in scale estimation]{\label{Fig.sub.4}
		\includegraphics[width=0.2\textwidth]{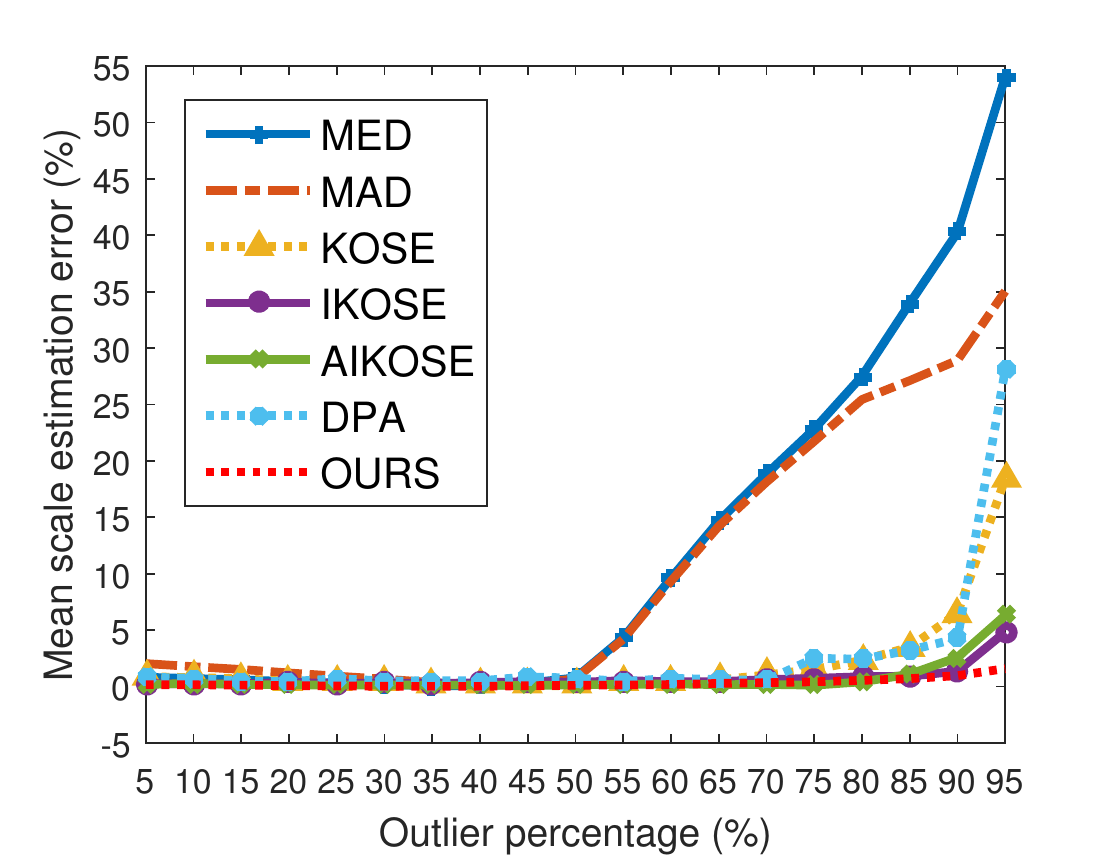}}
	\subfigure[The maximum errors in scale estimation]{\label{Fig.sub.4}
		\includegraphics[width=0.2\textwidth]{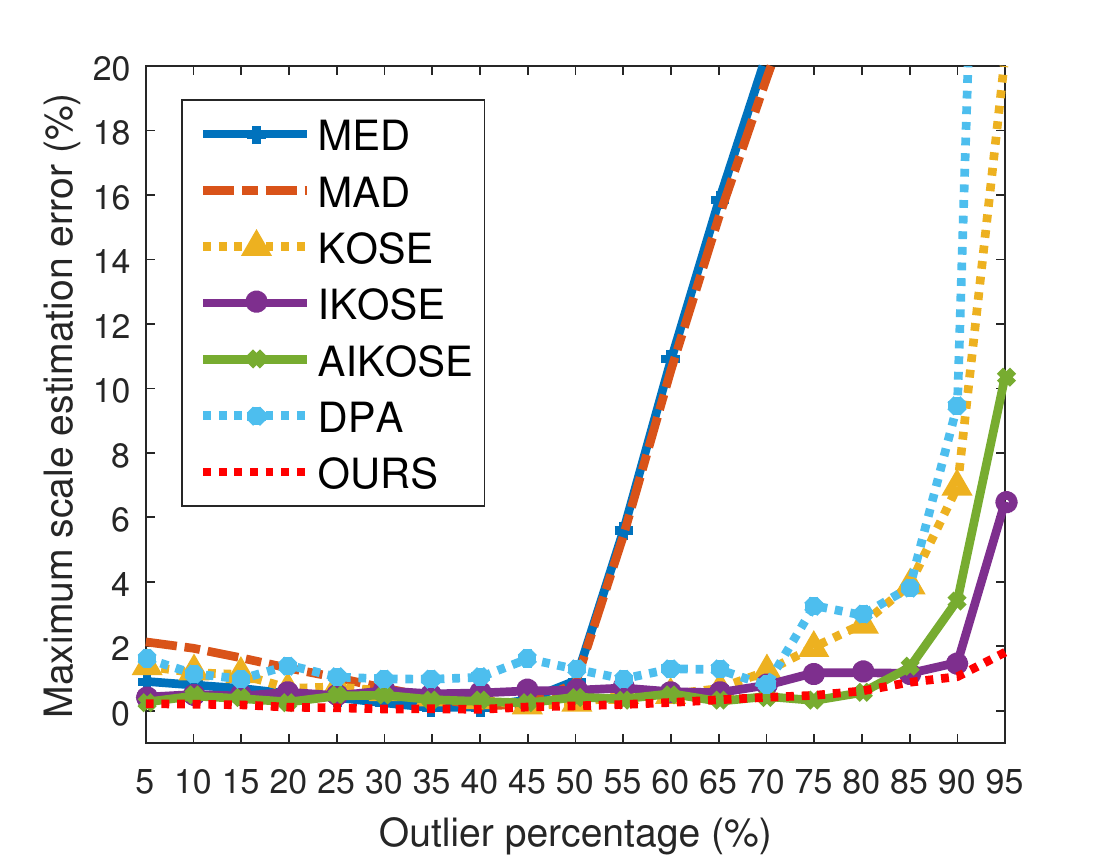}}
	\caption{Comparisons of the performance obtained by seven methods for scale estimation on synthetic data with 5\%-95\% outliers. (a) and (b) are respectively the data points with the outlier percentages on 50\% and 95\%. (c) and (d) display the mean and maximum errors among the scale estimation, which are obtained from all the competing methods. }\label{figuge.scaleestimation}
\end{figure}
%=====================================
\subsection{Experiments on Scale Estimation (Synthetic Data)}
\label{ssec:c40}
The experiments undertaken in this section are described as follows.
Two intersecting lines are generated lying on a plane that contains a total number of 2000 data points.
The number of the data points of the left line is decreased from 1900 to 100, which means that the outlier ratio is gradually increased from 5\% to 95\%. Meanwhile, the right line is hold fixed at 100 data points.
%==============================================================
We report the standard variances, the mean scale estimation errors, the median scale estimation errors and the maximum scale estimation errors in Table \ref{tab:quantitative}. We then respectively display the data points with the outlier percentages on 50\% and 95\% in Fig. \ref{figuge.scaleestimation} (a) and \ref{figuge.scaleestimation} (b), and the mean and maximum errors in Fig. \ref{figuge.scaleestimation} (d) and \ref{figuge.scaleestimation} (e).
Similar to \cite{wang2012simultaneously}, we use Eq. (\ref{equ:bestscale_only}) to compute the scale of inlier noise.
%=============================================================
\begin{table}[!ht]
	\centering
	\caption{Quantitative evaluation of the seven inlier scale estimation methods on synthetic data (the best results are boldfaced).}
	\scalebox{0.8}{\tabcolsep0.07in
		\begin{tabular}{lrrrrrrr}
			\toprule
			& \multicolumn{1}{c}{MED} & \multicolumn{1}{c}{MAD} & \multicolumn{1}{c}{KOSE} & \multicolumn{1}{c}{IKOSE} & \multicolumn{1}{c}{AIKOSE} & \multicolumn{1}{c}{DPA} & \multicolumn{1}{c}{OURS} \\
			\midrule
			Std.  & 16.43  & 12.01  & 4.26  & 1.03  & 1.48  & 6.26  & \textbf{0.40 } \\
			Mean  & 12.13  & 10.19  & 2.04  & 0.74  & 0.68  & 2.59  & \textbf{0.32 } \\
			Med.  & 0.83  & 2.04  & 0.59  & 0.44  & 0.19  & 0.74  & \textbf{0.16 } \\
			Max.  & 54.02  & 35.04  & 18.40  & 4.84  & 6.32  & 28.05  & \textbf{1.59 } \\
			\bottomrule
	\end{tabular}}
	\label{tab:quantitative}
\end{table}
%===========================================================
	\begin{equation}
			\label{equ:bestscale_only}
			s^h=\left|\tilde{r}_{\kappa} \right| \Big/ \Phi^{-1}{\left( \frac{1+\kappa/\tilde{n}}{2} \right)}
\end{equation}
where $\tilde{n}$ is the number of significant data points $\mathbf{\vartheta}^*$, which are selected according to the entropy.
Then, the scale estimation is measured through the following \cite{wang2012simultaneously}:
	\begin{equation}
			\label{equ:e8}
			\mathbf{\Lambda}(s_e, s_t) =  \max \Big(\frac{s_e}{s_t}, \frac{s_t}{s_e} \Big)-1
	\end{equation}
where $s_t$ is the true scale, and $s_e$ is the estimated scale.
As displayed in Table \ref{tab:quantitative} and Fig. \ref{figuge.scaleestimation}, all the scale estimation methods can work well when the outlier ratio is less than 50\%.
However, MED and MAD fail to estimate the scales when the outlier ratio is larger than 50\%.
The error of KOSE begins to increase when the outlier ratio is larger than 65\%, but is still better than MED and MAD.
KOSE and DPA become gradually worse when the outlier ratio is larger than 75\%.
IKOSE and AIKOSE can also achieve better results than other methods when the outlier ratio is larger than 90\%.
Among all the competing methods, the proposed method is able to achieve the best results, since it can adaptively estimate the inlier scale.
%===========================================================
\begin{figure}[!t]
	\begin{center}
		\subfigure[Biscuitbookbox]{
			\begin{minipage}[b]{0.147\textwidth}
				\centerline{\includegraphics[height=0.7\textwidth,width=1.\textwidth]{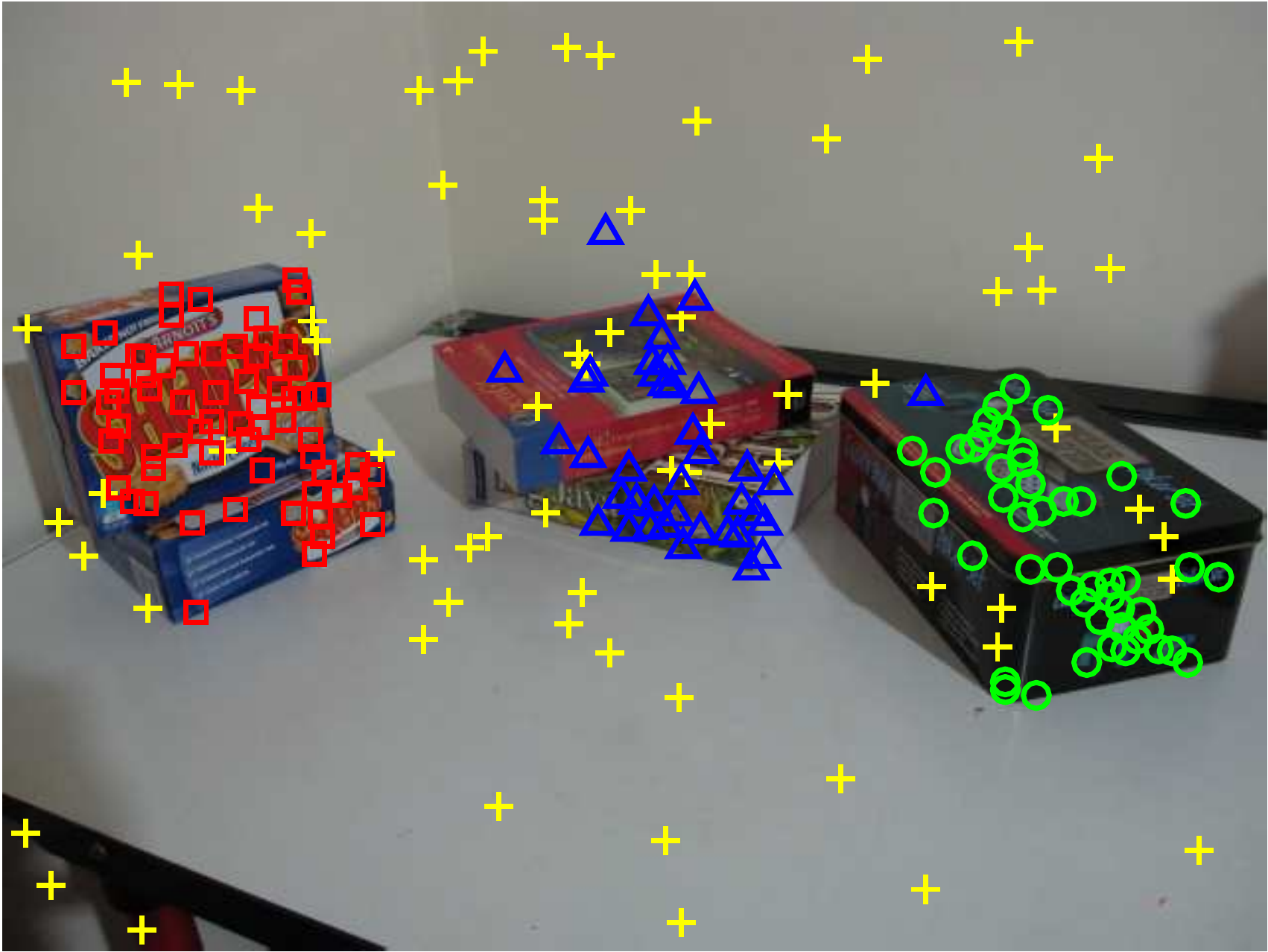}}
		\end{minipage}}
		\subfigure[Breadcartoychips]{
			\begin{minipage}[b]{0.147\textwidth}
				\centerline{\includegraphics[height=0.7\textwidth,width=1.\textwidth]{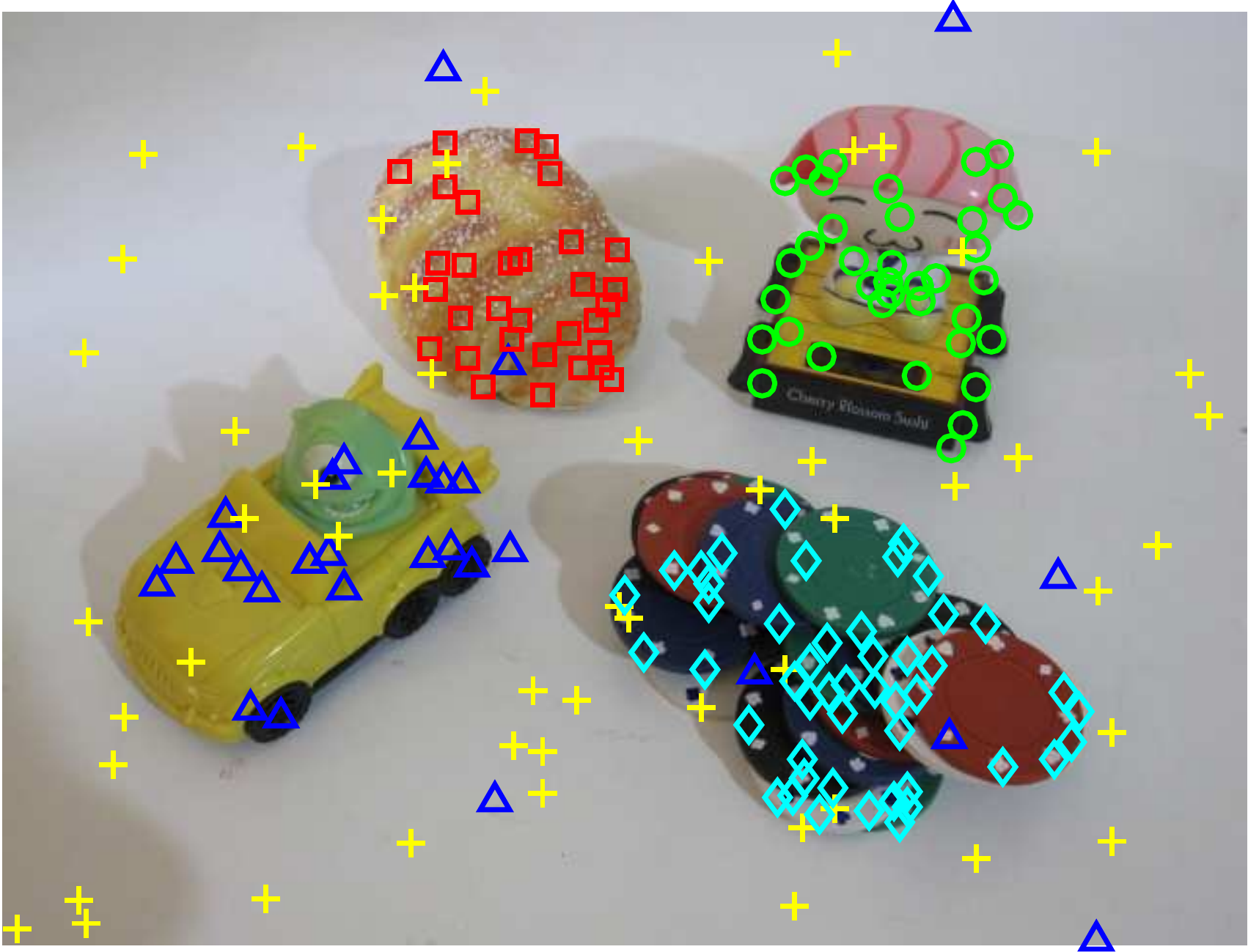}}
		\end{minipage}}
		\subfigure[Breadcubechips]{
			\begin{minipage}[b]{0.147\textwidth}
				\centerline{\includegraphics[height=0.7\textwidth,width=1.\textwidth]{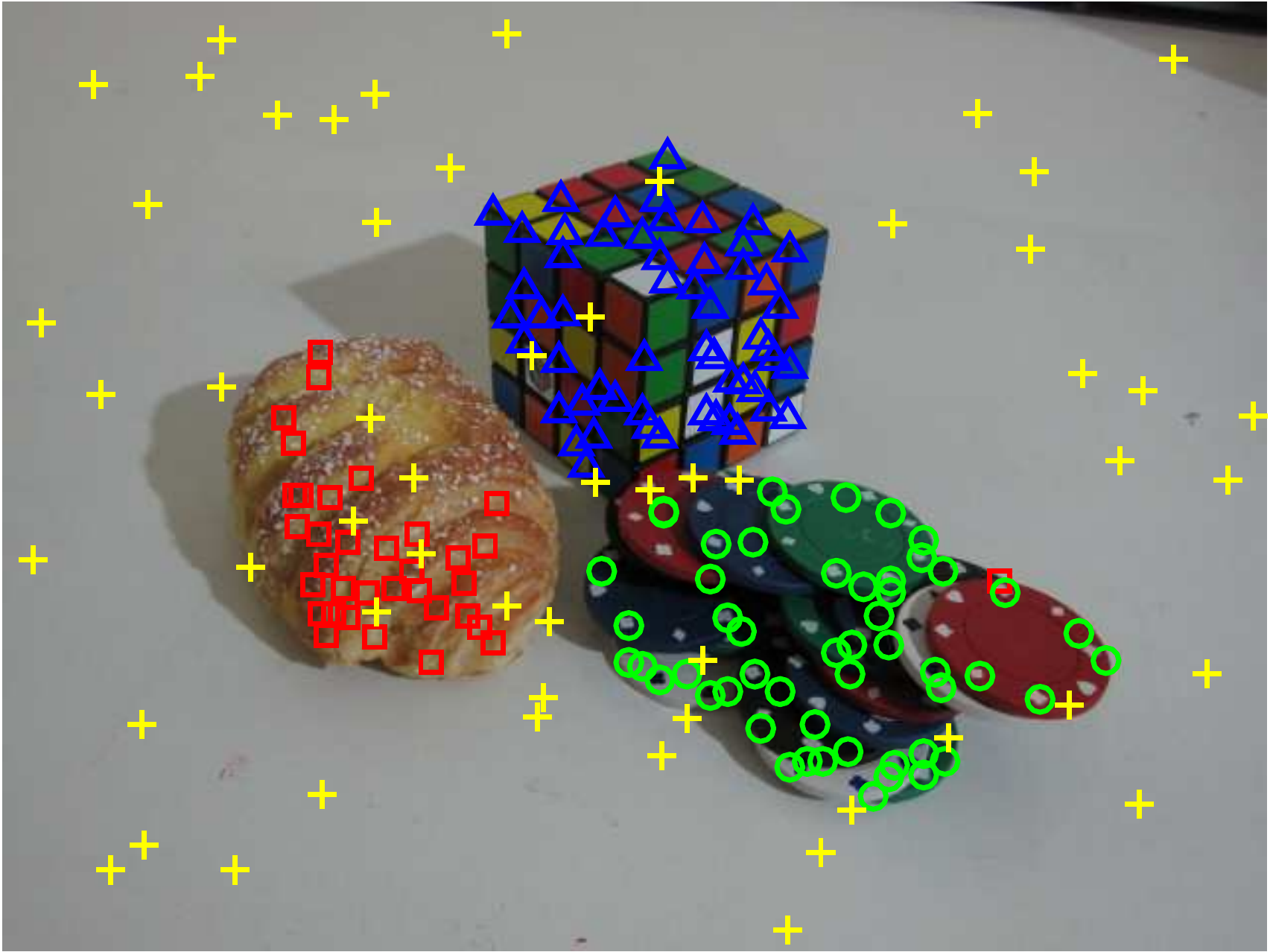}}
		\end{minipage}}
		\subfigure[Cubechips]{
			\begin{minipage}[b]{0.147\textwidth}
				\centerline{\includegraphics[height=0.7\textwidth,width=1.\textwidth]{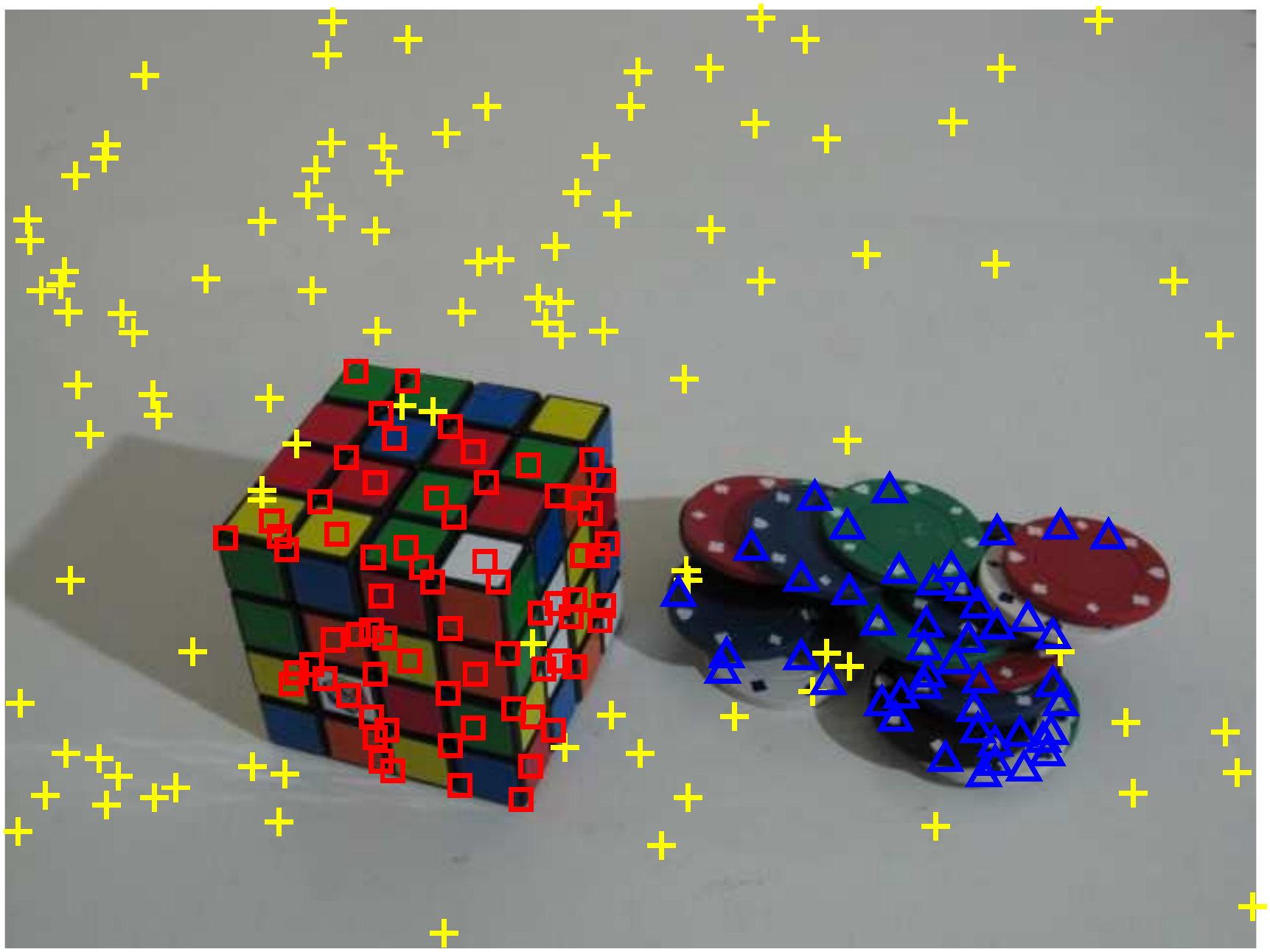}}
		\end{minipage}}
		\subfigure[Cubetoy]{
			\begin{minipage}[b]{0.147\textwidth}
				\centerline{\includegraphics[height=0.7\textwidth,width=1.\textwidth]{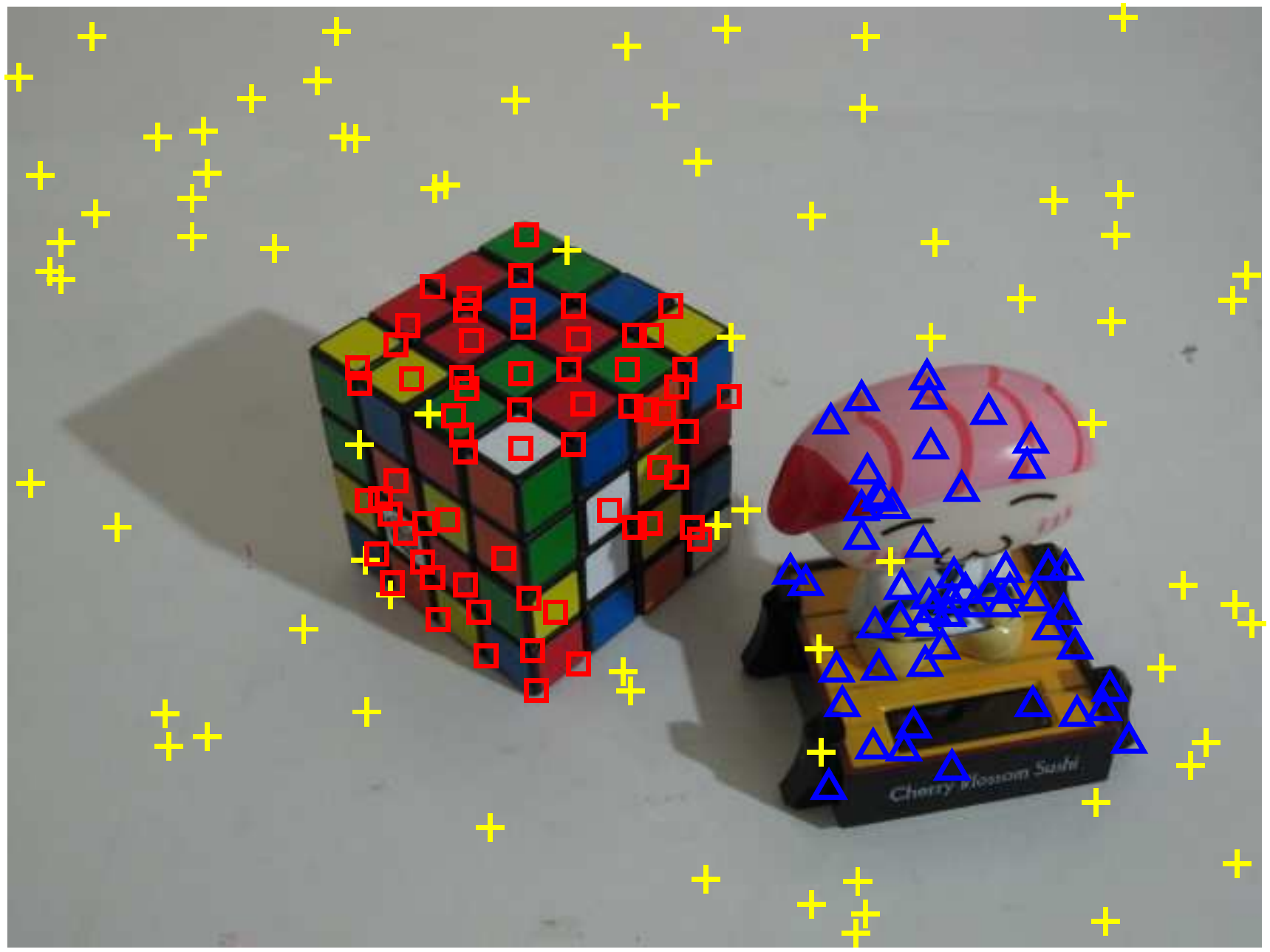}}
		\end{minipage}}
		\subfigure[Cube]{
			\begin{minipage}[b]{0.147\textwidth}
				\centerline{\includegraphics[height=0.7\textwidth,width=1.\textwidth]{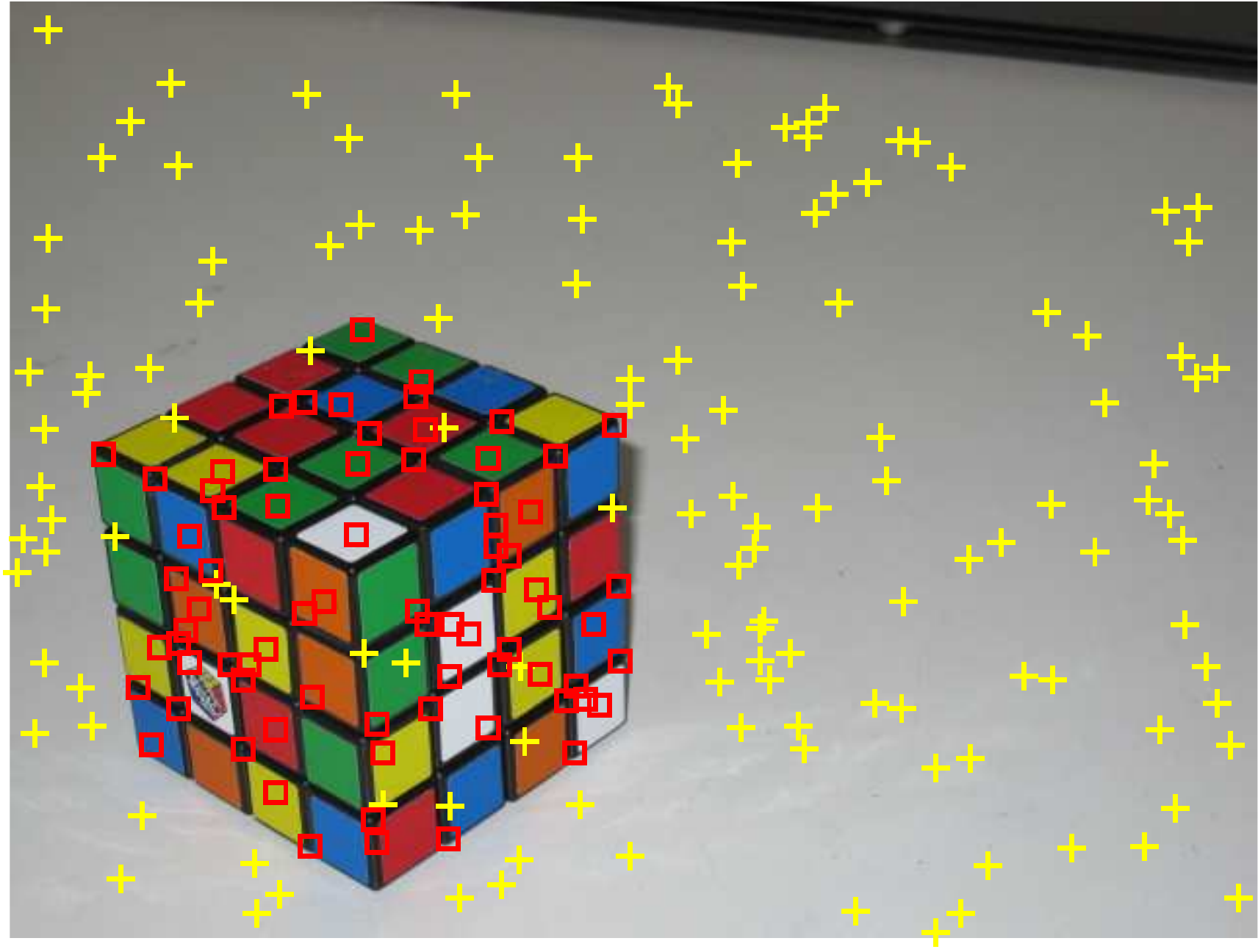}}
		\end{minipage}}
		\caption{Some results obtained by the proposed method  on six image pairs for two-view motion segmentation (only one view is shown).}
		\label{fig:fundamental}
	\end{center}
\end{figure}
%===========================================================
%==========================================================
\subsection{Experiments on Segmentation (Real Images)}
\label{ssec:c42}
In this section, we evaluate the performance of the six state-of-the-art model fitting methods on 16 representative image pairs with `single-structure' and `multiple-structural' data from the AdelaideRMF datasets \cite{wong2011dynamic}\footnote{http://cs.adelaide.edu.au/hwong/doku.php?id=data} for two-view motion segmentation and multi-homography segmentation, respectively.
Then, the average misclassification rates  and the CPU time (including sampling and fitting) are both reported in Table \ref{tab:fundamental} and Table \ref{tab:homography} for the two tasks, respectively.
The misclassification rate is adopted to measure the performance of these methods. It is defined as \cite{mittal2012generalized}:
	\begin{equation}
			\label{equ:error}
			\emph{error} =  \frac{\emph{number of misclassified points}}{\emph{total number of points}} \times 100\%.
	\end{equation}
\par
\begin{figure}[!t]
	\centering
	\includegraphics[width=0.265\textwidth]{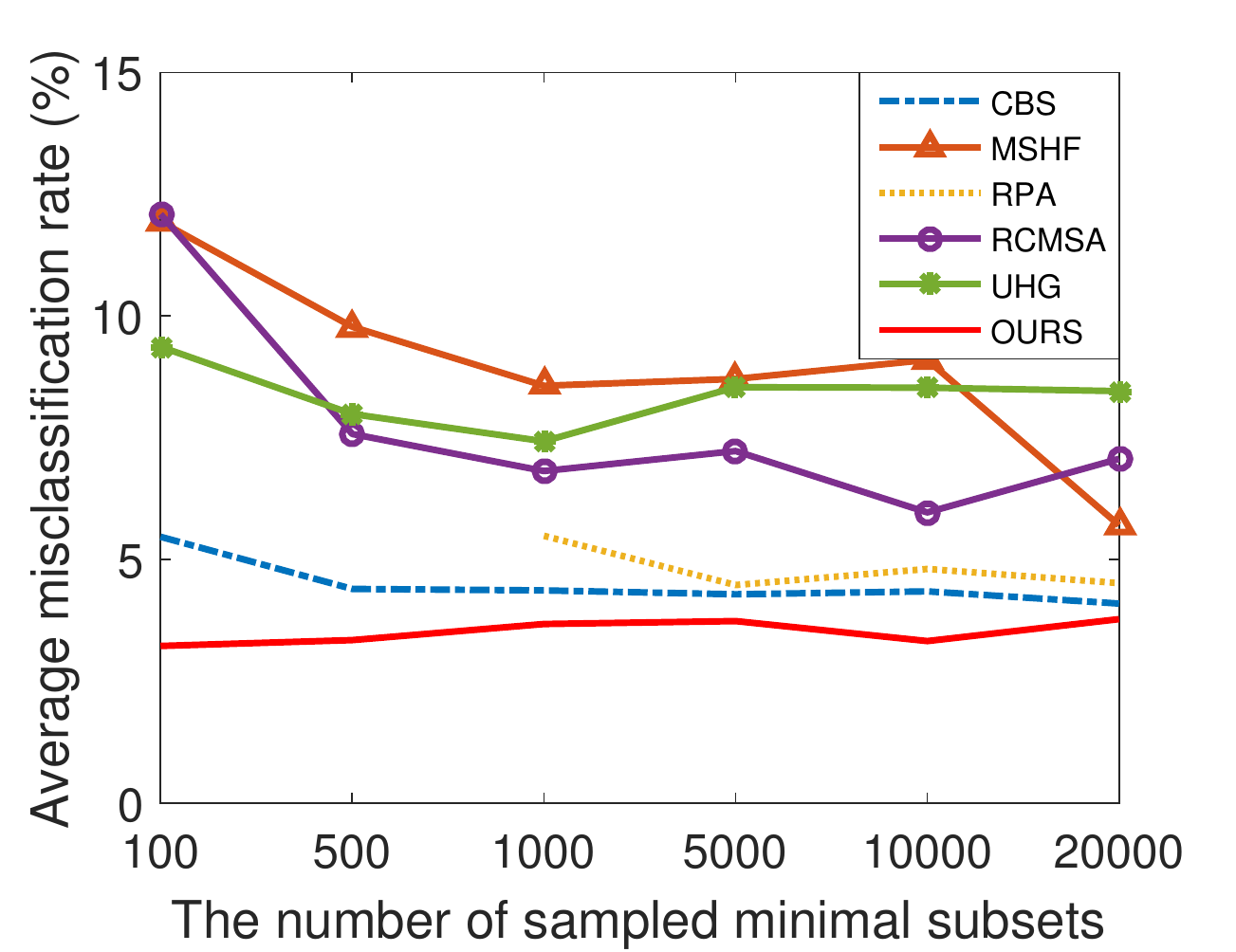}
	\caption{The average results are obtained by the six methods on a different number of sampled minimal subsets from the AdelaideRMF datasets.}\label{fig:sampling}
\end{figure}

The sampling frequency has a significant influence on sampling time. More sampled minimal subsets can usually achieve better segmentation results.
In fairness to the best accuracy of all the competing methods in our experiment, we also analyze the influence of sampling frequency on the six methods, where the number of minimal subsets is gradually increased from 100 to 20000 times on both fundamental matrix estimation and homography estimation.
We repeat the experiments 50 times and show the mean results in Fig. \ref{fig:sampling}.
Note that RPA fails to obtain the fitting results on a number of image pairs, when the sampling frequency is 100 and 500 times. Therefore, the results on these two sampling frequencies are not given.
As shown in Fig. \ref{fig:sampling}, the experimental results show that CBS, MSHF, RPA, RCMSA and UHG achieve the minimum average misclassification rates at 500, 20000, 5000, 10000 and 1000 times, respectively, and these values will be used in all experiments.
%===========================================================
%================================================
\begin{table}[!hb]
	\small
	\centering
	\caption{Misclassification rates (in percentage) and the CPU time (in seconds) for two-view motion segmentation on six methods (the best results are boldfaced).}
	\scalebox{0.8}{\tabcolsep0.08in
    \begin{tabular}{llrrrrrr}
    \toprule
    Data  &       & \multicolumn{1}{c}{M1} & \multicolumn{1}{c}{M2} & \multicolumn{1}{c}{M3} & \multicolumn{1}{c}{M4} & \multicolumn{1}{c}{M5} & \multicolumn{1}{c}{M6} \\
    \midrule
          & Std.  & \textbf{0.37} & 0.97  & 1.72  & 1.14  & 1.51  & 2.65  \\
          & Mean  & \textbf{0.39} & 3.51  & 4.53  & 4.96  & 5.14  & 5.18  \\
          & Med.  & \textbf{0.39} & 3.68  & 4.26  & 4.84  & 5.02  & 4.63  \\
    {Biscuitbookbox} & Time  & 3.90  & 26.64  & 122.56  & 105.76  & 10.16  & \textbf{2.66} \\
    \midrule
           & Std.  & 4.83  & 1.28  & \textbf{0.69} & 1.90  & 5.52  & 2.83  \\
           & Mean  & \textbf{5.76} & 27.84  & 8.70  & 9.18  & 10.76  & 6.09  \\
           & Med.  & \textbf{3.68} & 28.14  & 8.66  & 9.31  & 8.02  & 5.70  \\
     {Breadcartoychips} & Time  & 3.36  & 28.79  & 115.47  & 93.88  & 9.19  & \textbf{2.73} \\
     \midrule
          & Std.  & \textbf{1.16} & 1.92  & 2.84  & 4.47  & 7.80  & 3.99  \\
          & Mean  & \textbf{2.48} & 5.04  & 5.57  & 10.87  & 9.04  & 4.50  \\
          & Med.  & \textbf{2.39} & 5.65  & 6.09  & 9.13  & 4.35  & 4.87  \\
    {Breadcubechips} & Time  & 3.61  & 24.77  & 108.55  & 91.90  & 8.99  & \textbf{1.82} \\
    \midrule
           & Std.  & \textbf{0.11} & 0.84  & 0.56  & 3.65  & 0.65  & 0.33  \\
           & Mean  & \textbf{0.71} & 2.64  & 2.92  & 5.52  & 3.28  & 1.01  \\
           & Med.  & \textbf{0.68} & 2.37  & 3.05  & 5.19  & 3.31  & 0.99  \\
     {Cube } & Time  & 12.87  & 30.57  & 129.35  & 177.21  & 11.92  & \textbf{3.20} \\
     \midrule
          & Std.  & 0.94  & \textbf{0.68} & 1.18  & 5.75  & 6.90  & 0.86  \\
          & Mean  & 1.94  & 15.00  & 5.38  & 12.52  & 19.08  & \textbf{1.49} \\
          & Med.  & 1.59  & 15.13  & 5.41  & 11.31  & 14.98  & \textbf{1.53} \\
    {Cubebreadtoychips} & Time  & 4.21  & 28.66  & 169.13  & 138.08  & 12.63  & \textbf{3.85} \\
    \midrule
           & Std.  & 1.81  & 1.44  & 1.08  & 6.20  & 4.40  & \textbf{0.38} \\
           & Mean  & 2.17  & 3.79  & 4.30  & 7.40  & 6.69  & \textbf{0.25} \\
           & Med.  & 1.26  & 3.61  & 4.15  & 5.05  & 5.11  & \textbf{0.00} \\
     {Cubechips} & Time  & 6.05  & 33.16  & 134.04  & 117.55  & 11.15  & \textbf{3.22} \\
     \midrule
          & Std.  & 0.81  & 2.33  & \textbf{0.71} & 2.17  & 1.10  & 1.64  \\
          & Mean  & \textbf{1.13} & 5.86  & 3.64  & 6.40  & 4.54  & 1.90  \\
          & Med.  & \textbf{1.05} & 5.44  & 3.56  & 6.90  & 5.02  & 1.61  \\
    {Cubetoy} & Time  & 4.96  & 31.79  & 97.72  & 96.01  & 9.56  & \textbf{2.92} \\
    \midrule
           & Std.  & \textbf{0.21} & 1.88  & 1.01  & 5.71  & 0.76  & 0.60  \\
           & Mean  & \textbf{0.13} & 2.70  & 3.30  & 8.87  & 3.43  & 0.56  \\
           & Med.  & \textbf{0.00} & 2.61  & 3.04  & 5.43  & 3.43  & 0.43  \\
     {Game } & Time  & 9.02  & 22.34  & 100.70  & 153.58  & 9.05  & \textbf{2.24} \\
    \midrule
          & Std.  & 1.28  & 1.42  & \textbf{1.22} & 3.87  & 3.58  & 1.66  \\
          & Mean  & \textbf{1.84} & 8.30  & 4.79  & 8.21  & 7.75  & 2.62  \\
          & Med.  & \textbf{1.38} & 8.33  & 4.78  & 7.15  & 6.15  & 2.47  \\
    {Average} & Time  & 6.00  & 28.34  & 122.19  & 121.75  & 10.33  & \textbf{2.83} \\
    \bottomrule
	\end{tabular}}%
	\medskip
	\raggedright{
		\scriptsize{(M1-CBS; M2-MSHF; M3-RPA; M4-RCMSA; M5-UHG; M6-HOMF.)}}
	\label{tab:fundamental}%
\end{table}%
%============================================
The proposed method obtains the stable average misclassification rates due to the IHO, the AIE and the guided sampling. Therefore, we fix the sampling frequency to 200 times for the proposed method in the experiments.
%===========================================================
%---------------------------------------------
\begin{table}[!hb]
	\small
	\centering
	\caption{Misclassification rates (in percentage) and the CPU time (in seconds) for multi-homography segmentation on six methods (the best results are boldfaced).}
	\scalebox{0.8}{\tabcolsep0.08in
    \begin{tabular}{llrrrrrr}
    \toprule
    Data  &       & \multicolumn{1}{c}{M1} & \multicolumn{1}{c}{M2} & \multicolumn{1}{c}{M3} & \multicolumn{1}{c}{M4} & \multicolumn{1}{c}{M5} & \multicolumn{1}{c}{M6} \\
    \midrule
          & Std.  & 0.18  & 1.40  & 5.70  & 1.35  & 0.24  & \textbf{0.14} \\
          & Mean  & 0.08  & 1.39  & 4.09  & 2.64  & 2.37  & \textbf{0.05} \\
          & Med.  & \textbf{0.00} & 1.36  & 2.07  & 2.59  & 2.53  & \textbf{0.00} \\
    {Bonython} & Time  & 4.75  & 11.93  & 103.61  & 156.50  & 28.46  & \textbf{1.29} \\ 
    \midrule
           & Std.  & \textbf{0.00} & 1.12  & 0.00  & 6.20  & 6.43  & 0.07  \\
           & Mean  & \textbf{0.47} & 1.97  & 1.40  & 8.21  & 15.14  & 0.93  \\
           & Med.  & \textbf{0.47} & 2.15  & 1.40  & 6.07  & 14.25  & 0.93  \\
     {Elderhalla} & Time  & 4.24  & 14.74  & 128.02  & 113.89  & 28.86  & \textbf{1.27} \\
     \midrule
          & Std.  & 2.03  & 1.77  & 4.08  & 6.32  & \textbf{0.54} & 1.63  \\
          & Mean  & 12.72  & 4.42  & 12.07  & 7.17  & \textbf{3.40} & 11.94  \\
          & Med.  & 13.17  & 4.25  & 11.05  & 5.67  & \textbf{3.22} & 12.33  \\
    {Johnsona} & Time  & 4.30  & 27.25  & 240.34  & 165.64  & 34.13  & \textbf{1.71} \\
    \midrule
           & Std.  & 16.97  & 5.00  & 0.90  & 1.74  & \textbf{0.85} & 13.49  \\
           & Mean  & 24.17  & \textbf{3.70} & 7.13  & 7.48  & 7.59  & 16.38  \\
           & Med.  & 26.74  & \textbf{1.96} & 7.17  & 7.39  & 7.47  & 5.60  \\
     {Neem } & Time  & 4.07  & 21.12  & 150.43  & 91.85  & 29.60  & \textbf{1.32} \\
     \midrule
          & Std.  & 0.35  & 3.77  & 0.46  & \textbf{0.00} & 2.17  & 0.50  \\
          & Mean  & 0.62  & 2.70  & 2.12  & 0.83  & 2.60  & \textbf{0.46} \\
          & Med.  & 0.41  & 0.62  & 2.07  & 0.83  & 1.97  & \textbf{0.39} \\
    {Nese } & Time  & 4.12  & 22.88  & 136.90  & 98.37  & 29.67  & \textbf{1.29} \\
    \midrule
           & Std.  & 0.49  & 0.53  & 0.23  & \textbf{0.00} & 0.22  & 10.71  \\
           & Mean  & 3.97  & 1.13  & 2.67  & \textbf{0.55} & 19.13  & 10.12  \\
           & Med.  & 3.99  & 1.24  & 2.75  & \textbf{0.55} & 19.00  & 2.90  \\
     {Oldclassicswing} & Time  & 4.61  & 43.93  & 228.49  & 178.88  & 32.60  & \textbf{1.76} \\
     \midrule
          & Std.  & 0.40  & 0.61  & 0.18  & \textbf{0.00} & 7.51  & 0.46  \\
          & Mean  & 0.85  & 0.97  & 2.20  & 1.27  & 5.16  & \textbf{0.42} \\
          & Med.  & 0.85  & \textbf{0.80} & 2.12  & 1.27  & 1.60  & 0.80  \\
    {Sene } & Time  & 4.09  & 20.48  & 150.17  & 100.57  & 29.05  & \textbf{1.31} \\
    \midrule
           & Std.  & 0.16  & 0.20  & 0.17  & 0.40  & \textbf{0.10} & 1.50  \\
           & Mean  & 0.87  & \textbf{0.38} & 1.61  & 1.68  & 1.54  & 1.30  \\
           & Med.  & 0.93  & \textbf{0.31} & 1.56  & 1.56  & 1.51  & 0.40  \\
     {Unionhouse} & Time  & 7.45  & 20.40  & 199.34  & 299.50  & 33.20  & \textbf{1.55} \\
    \midrule
          & Std.  & 2.57  & 1.80  & \textbf{1.46} & 2.00  & 2.26  & 3.56  \\
          & Mean  & 5.47  & \textbf{2.08} & 4.16  & 3.73  & 7.12  & 5.20  \\
          & Med.  & 5.82  & \textbf{1.59} & 3.78  & 3.24  & 6.44  & 2.92  \\
    {Average} & Time  & 4.70  & 22.84  & 167.16  & 150.65  & 30.70  & \textbf{1.44} \\
    \bottomrule
	\end{tabular}}%
	\medskip
	\raggedright{
		\scriptsize{(M1-CBS; M2-MSHF; M3-RPA; M4-RCMSA; M5-UHG; M6-HOMF.)}}
	\label{tab:homography}%
\end{table}%
%=====================================================
%=====================================================
\begin{figure}[ht]
	\begin{center}
		\subfigure[Bonython]{
			\begin{minipage}[b]{0.147\textwidth}
				\centerline{\includegraphics[height=0.7\textwidth,width=1.\textwidth]{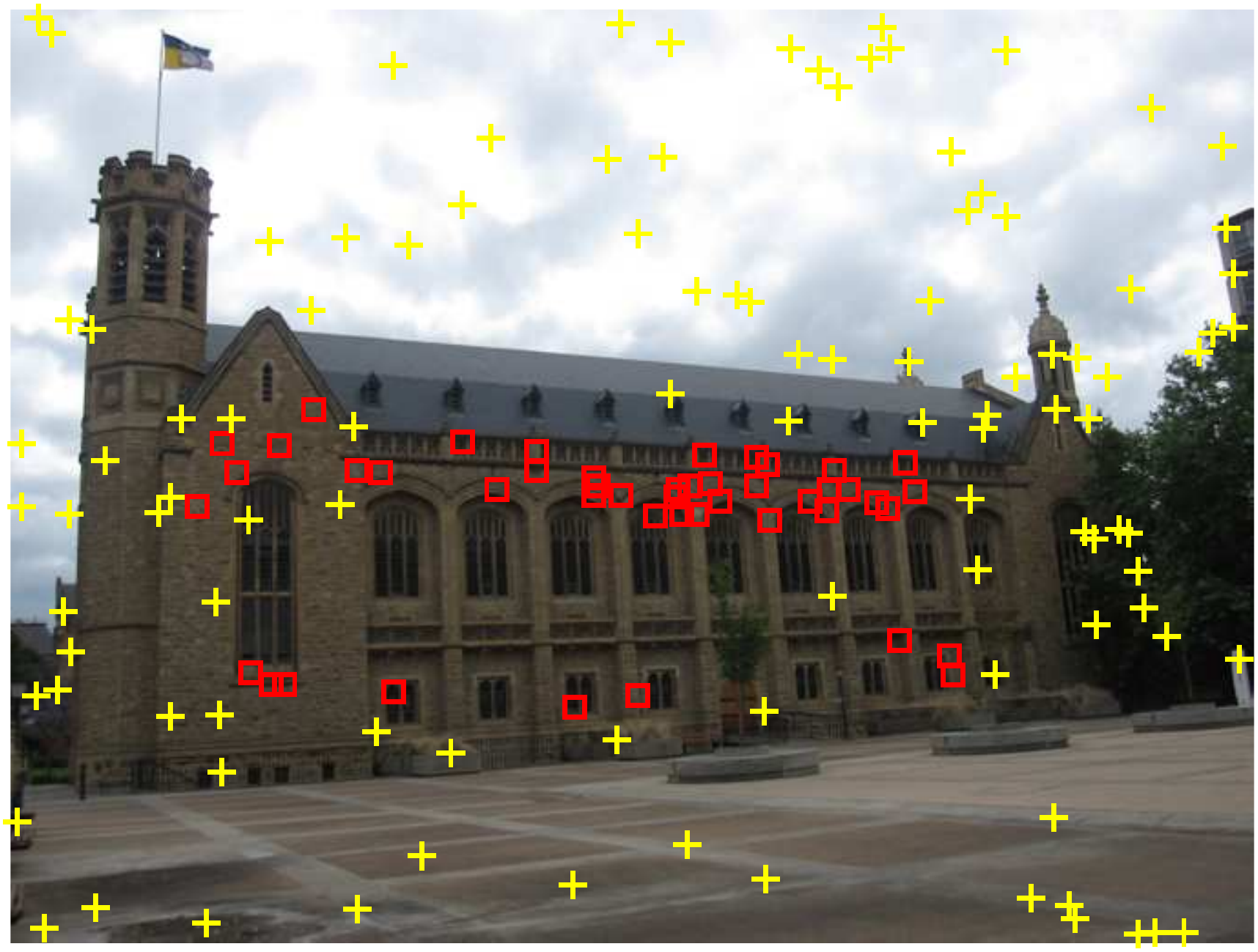}}
		\end{minipage}}
		\subfigure[Elderhalla]{
			\begin{minipage}[b]{0.147\textwidth}
				\centerline{\includegraphics[height=0.7\textwidth,width=1.\textwidth]{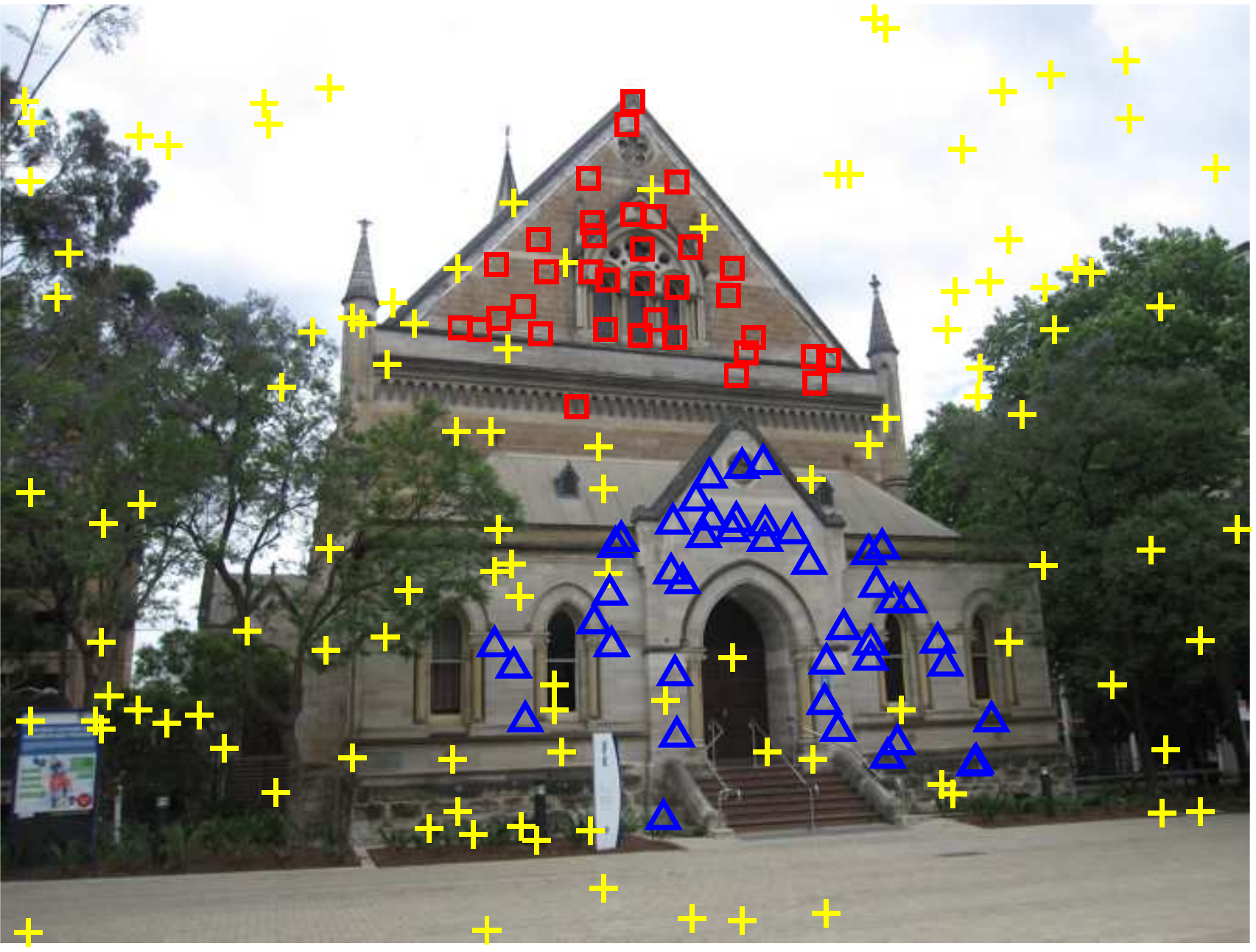}}
		\end{minipage}}
		\subfigure[Johnsona]{
			\begin{minipage}[b]{0.147\textwidth}
				\centerline{\includegraphics[height=0.7\textwidth,width=1.\textwidth]{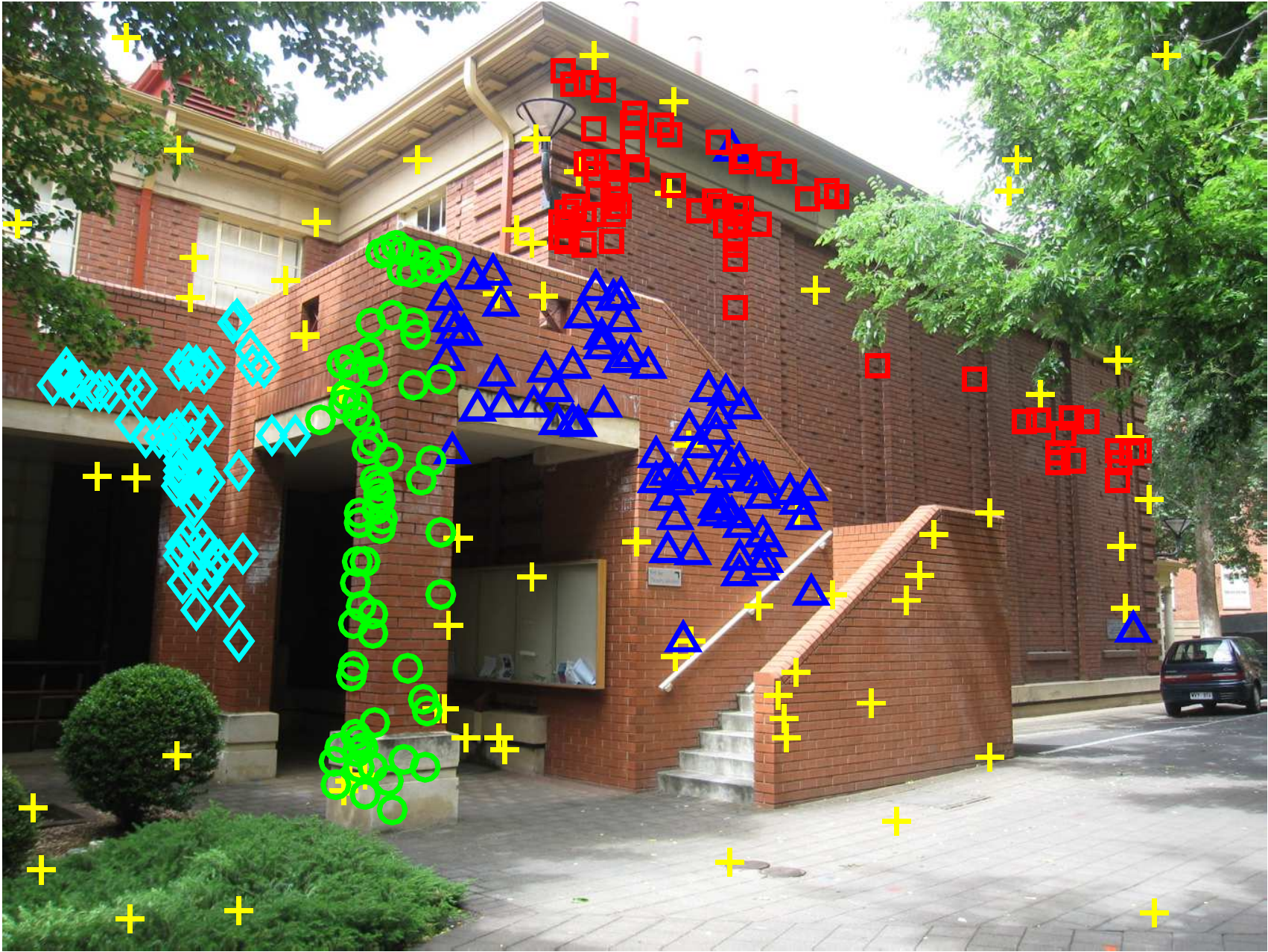}}
		\end{minipage}}
		\subfigure[Neem]{
			\begin{minipage}[b]{0.147\textwidth}
				\centerline{\includegraphics[height=0.7\textwidth,width=1.\textwidth]{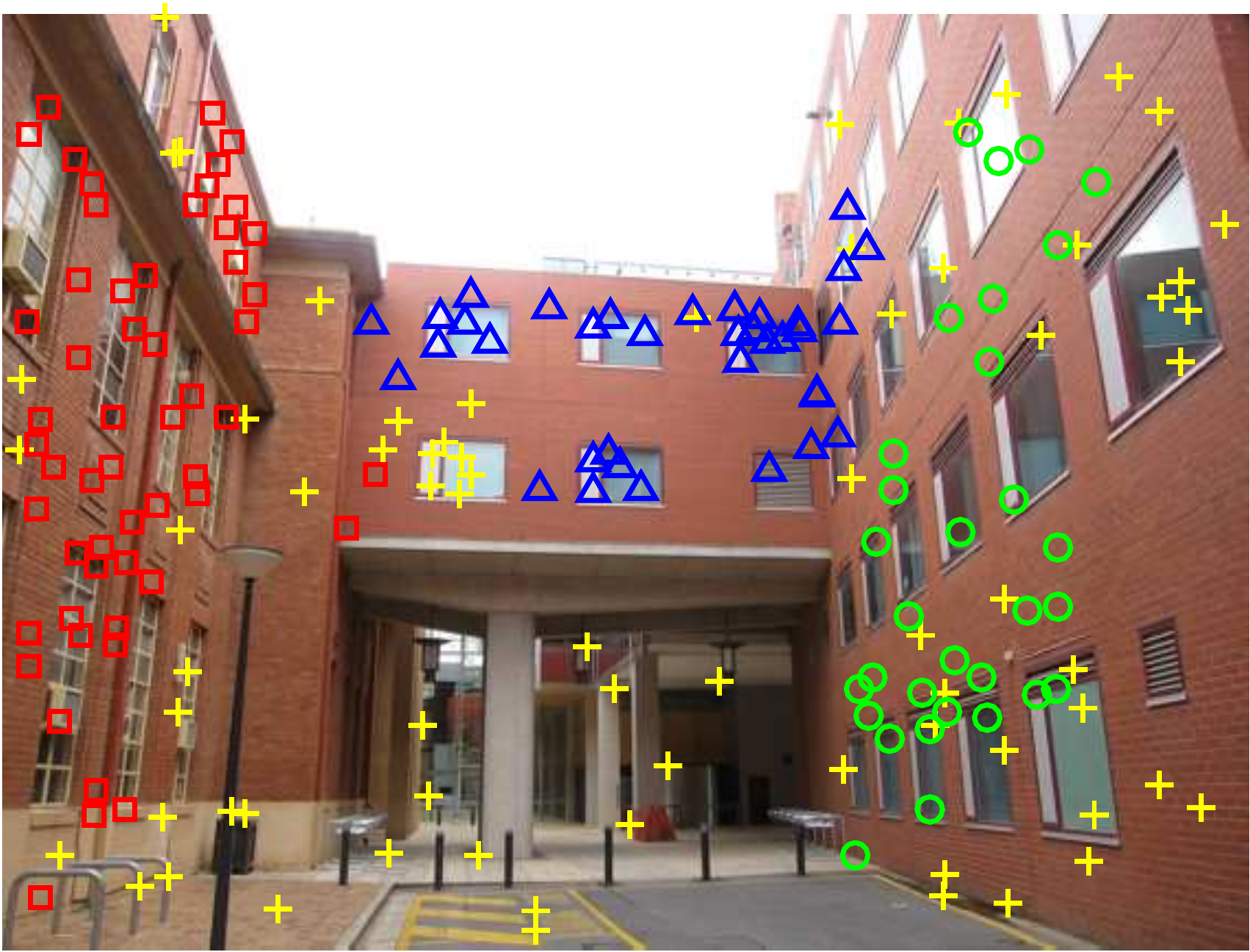}}
		\end{minipage}}
		\subfigure[Nese]{
			\begin{minipage}[b]{0.147\textwidth}
				\centerline{\includegraphics[height=0.7\textwidth,width=1.\textwidth]{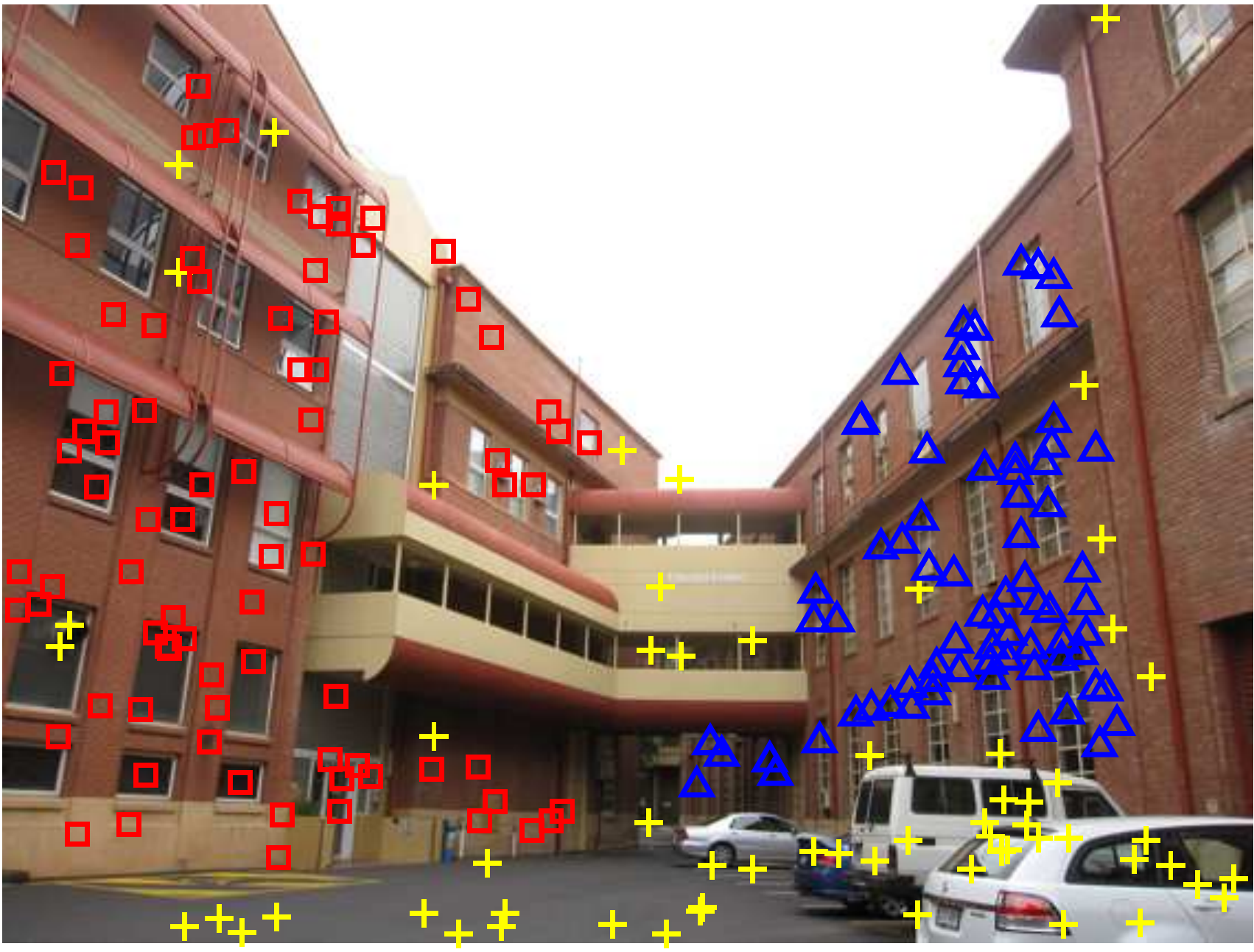}}
		\end{minipage}}
		\subfigure[Sene]{
			\begin{minipage}[b]{0.147\textwidth}
				\centerline{\includegraphics[height=0.7\textwidth,width=1.\textwidth]{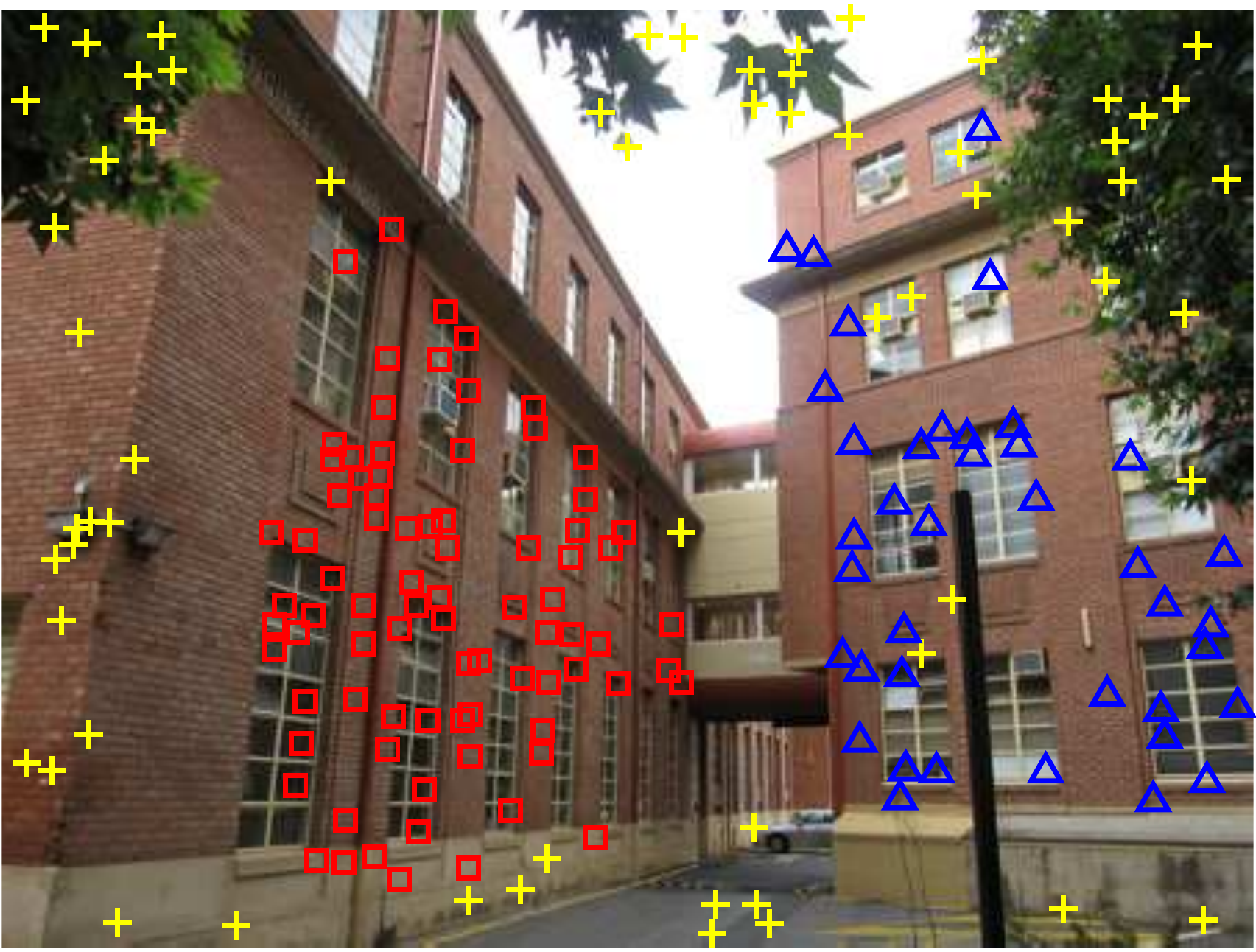}}
		\end{minipage}}
		\caption{ Some results obtained by the proposed method  on six image pairs for multi-homography segmentation (only one view is shown).}
		\label{fig:homography}
	\end{center}
\end{figure}
%====================================================
\subsubsection{Two-view Motion Segmentation}
\label{ssec:c422}
In this section, we evaluate the partitioning capability of the six competing methods to identify two-view motion segmentation.
From the data reported in Table \ref{tab:fundamental} and Fig. \ref{fig:fundamental} (except for Cubebreadtoychips and Game due to the space limit), we can see that our method achieves the fastest running time (in seconds) among all the competing methods.
Although the segmentation accuracy of HOMF is slightly lower than CBS, it has significantly improved computational speed over CBS for all the representative image pairs.
CBS achieves the lowest average misclassification rates due to the $k$th-order cost function and the greedy algorithm.
RPA achieves the third lowest average misclassification rates, however it takes more time to sample the minimal subset than other methods.
UHG achieves relatively good performance due to the promising hypothesis generation which is effectively accelerated.
RCMSA and MSHF obtain similar average misclassification rates, but MSHF runs much faster than RCMSA. This is because that RCMSA employs a simulated annealing framework, which is time-consuming.
In contrast, our method achieves good average misclassification rates (only slightly worse than CBS) with low computational cost. Therefore, our method achieves good tradeoff between the segmentation performance and running time.
%===================================================================
\subsubsection{Multi-homography Segmentation}
\label{ssec:c421}
In this section, we also evaluate the performance of the six competing methods for multi-homography segmentation.
From the data reported in Table \ref{tab:homography} and Fig. \ref{fig:homography} (omitting Oldclassicswing and Unionhouse due to the space limit), we can see that our method can more efficiently recover the real plane structure on multi-homography segmentation.
Our method achieves superior speed (in seconds) over the other competing methods, although the average misclassification rates are higher than MSHF and RCMSA. This is because that the lower number of sampled minimal subsets leads to failure of multi-structural data with high complexity (e.g., Fig. \ref{fig:homography} (c)).
MSHF achieves the lowest average misclassification rate (in percentage) among all the competing methods because of the effectiveness of the constructed hypergraph. However, the running time of MSHF is slower than our method.
Both RCMSA and RPA achieve similar results in accuracy and performance, but obtain slow speeds due to the time-consuming sampling process.
CBS fails in the image pairs due to the loss of useful information during the data sub-sampling strategy.
UHG does not achieve reliable fitting performance since it selects the model instance by using T-Linkage \cite{magri2014t}.
Nevertheless, the experimental results show that HOMF performs faster than the other five competing methods in practice, including sampling and fitting time.
Experimental results show that our method can segment multi-structural data with outliers quickly and efficiently.

\section{Conclusion}
\label{sec:Conclusion}
In this paper, we have developed a novel hypergraph optimization based model fitting (HOMF) method, which aims to rapidly fit multi-structural data contaminated with a large number of noise and outlier data points. We construct a simple but effective hypergrah based on the generated model hypotheses.
In the constructed hypergraph, each vertex represents a data point and each hyperedge denotes a model hypothesis.
To optimize the hypergraph, we develop the AIE scale estimator and the IHO hyperedge optimization algorithm for optimizing the hypergraph.
In particular, we efficiently generate an optimized hyperedge by IHO, and then employ AIE to distinguish significant vertices from insignificant vertices.
The significant vertices are used to construct the optimized hypergraph for reducing the computational complexity, and the insignificant vertices are used to guide the following sampling for different structures.
Based on the constructed hypergraph, the hyperedges and the vertices can be effectively optimized during each iteration.
The two parts are tightly coupled to optimize the hypergraph for both computational efficiency and accuracy.
The experimental results on synthetic data and real images have shown that the proposed HOMF method can obtain better performance and much faster than the other competing methods.
\subsubsection{Acknowledgments.}
We would like to thank anonymous reviewers for their suggestions. 
This work was supported by the National Natural Science Foundation of China under Grants U1605252, 61702431, 61472334, 61571379 and 61872307. David Suter acknowledges funding under ARC DP160103490.
\bibliographystyle{aaai}
\bibliography{swin}

\end{document}